\documentclass{article}

    \usepackage[final]{neurips_2025}

\usepackage[utf8]{inputenc} % allow utf-8 input
\usepackage[T1]{fontenc}    % use 8-bit T1 fonts
\usepackage{hyperref}       % hyperlinks
\usepackage{graphicx} % for \includegraphics
\usepackage{url}            % simple URL typesetting
\usepackage{booktabs}       % professional-quality tables
\usepackage{amsfonts}       % blackboard math symbols
\usepackage{nicefrac}       % compact symbols for 1/2, etc.
\usepackage{microtype}      % microtypography
\usepackage{xcolor}         % colors
\usepackage{amsmath}
\usepackage{cleveref}
\usepackage{multirow}
\usepackage{multicol}
\usepackage{makecell}

\title{RLGF: Reinforcement Learning with Geometric Feedback for Autonomous Driving Video Generation}

\author{
    \makecell[c]{
                    Tianyi Yan$^{1 *}$ \textbf{,}
                 Wencheng Han$^{1 *}$  \textbf{,}
                 Xia Zhou$^{2}$ \textbf{,}
                 Xueyang Zhang$^{2}$ \\
                 Kun Zhan$^{2}$ \textbf{,}
                 Cheng-zhong Xu$^{1}$ \textbf{,}
                 Jianbing Shen$^{1}$\textsuperscript{\dag}
                 }\\
    $^{1}$SKL-IOTSC, Computer and Information Science, University of Macau\textbf{,}
    $^{2}$Li Auto Inc. \\
}

\begin{document}

\maketitle

\footnotetext{
\small
\textsuperscript{*} Equal contribution, 
\textsuperscript{$\dagger$} Corresponding author,
This work was supported by the Science and Technology Development Fund of Macau SAR (FDCT) under grants 0102/2023/RIA2 and 0154/2022/A3 and 001/2024/SKL, and the Jiangyin Hi-tech Industrial Development Zone under the Taihu Innovation Scheme (EF2025-00003-SKL-IOTSC).
}

\begin{abstract}
Synthetic data is crucial for advancing autonomous driving (AD) systems, yet current state-of-the-art video generation models, despite their visual realism, suffer from subtle geometric distortions that limit their utility for downstream perception tasks. 
We identify and quantify this critical issue, demonstrating a significant performance gap in 3D object detection when using synthetic versus real data.
To address this, we introduce Reinforcement Learning with Geometric Feedback (RLGF), RLGF uniquely refines video diffusion models by incorporating rewards from specialized latent-space AD perception models. 
Its core components include an efficient Latent-Space Windowing Optimization technique for targeted feedback during diffusion, and a Hierarchical Geometric Reward (HGR) system providing multi-level rewards for point-line-plane alignment, and scene occupancy coherence. 
To quantify these distortions, we propose GeoScores. Applied to models like DiVE on nuScenes, RLGF substantially reduces geometric errors (e.g., VP error by 21\%, Depth error by 57\%) and dramatically improves 3D object detection mAP by 12.7\%, narrowing the gap to real-data performance. RLGF offers a plug-and-play solution for generating geometrically sound and reliable synthetic videos for AD development.

\end{abstract}

\begin{figure}
    \centering
    \includegraphics[width=\linewidth]{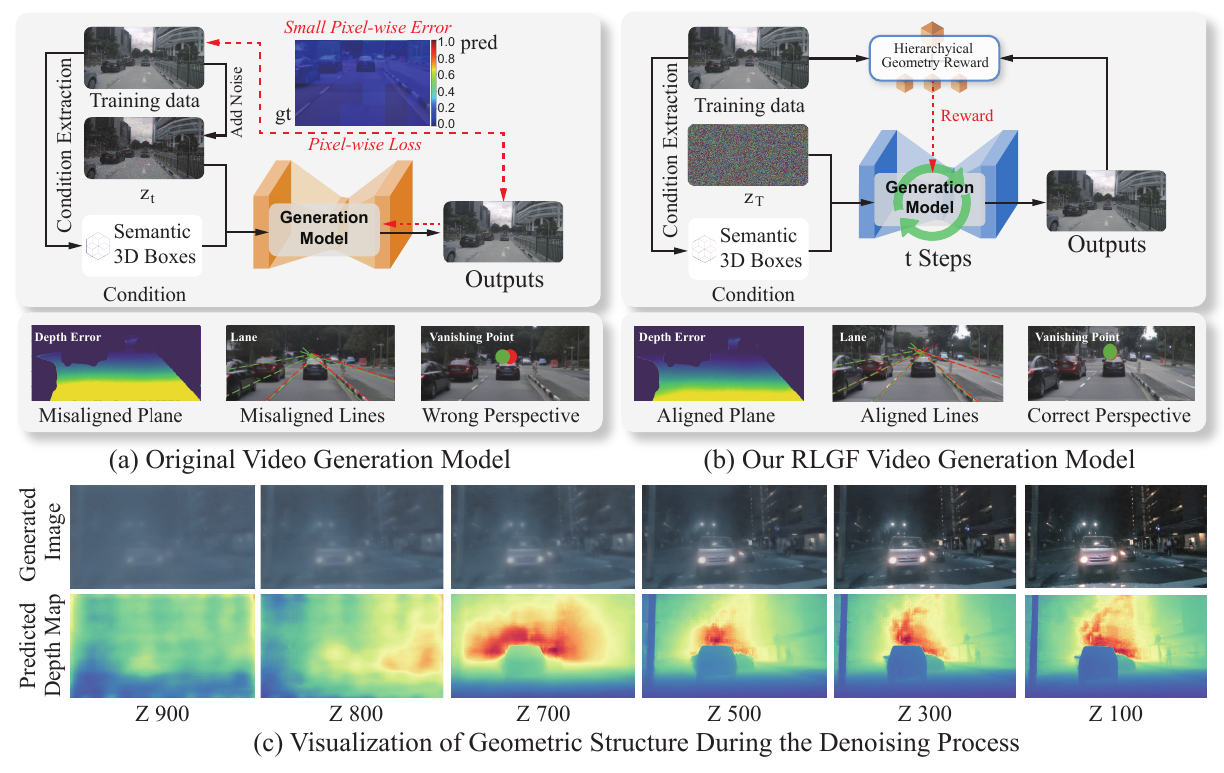}
    \vspace{-7mm}
    \caption{
    % \textbf{Overview of the geometric distortion problem and our proposed Reinforcement Learning with Geometric Feedback (RLGF) framework.} (a) Standard conditional video generation models often produce visually plausible results from noisy latents and conditions. (b) Generated videos often exhibit geometric distortions (e.g., VP shifts, depth errors) that severely degrade 3D object detection (e.g., mAP drop from 35.5 to 25.7). (c) Our RLGF framework injects geometric feedback directly into the diffusion process. 
(a) \textbf{Original Video Generation Models}, optimized via pixel-level supervision (e.g., noise prediction error), often produce visually plausible videos that nonetheless suffer from severe geometric flaws (misaligned planes/lines, wrong perspective). This can degrade downstream tasks like 3D object detection (e.g., mAP drop from 35.5 to 25.7).
(b) \textbf{Our RLGF} integrates a Hierarchical Geometry Reward directly into the multi-step denoising process. This reward, derived from perception models, guides the generation model to produce outputs with aligned planes, correct lane structures, and accurate perspective.
(c) \textbf{Visualized depth maps} from noisy latents at various denoising stages (from noisy to less noisy) show coarse geometry emerging early and details later. This motivates our Latent-Space Windowing Optimization for targeted intermediate rewards.
    }
    % (d) RLGF substantially boosts 3D object detection (mAP, NDS) and geometric fidelity (GeoScores) in the synthesized videos.}
        \vspace{-5mm}
    \label{fig:motivation}
\end{figure}

\section{Introduction}
% 合成ad video很重要，现有ad video 生成方法合成的非常逼真，但是忽略了几何偏差
%fvd高,普遍使用的检测性能差。-》2d可以，3d很明显。所以3d信息错=误
% 问题：系统性几何偏差导致合成数据难以利用，表现在视觉效果好，但是存在（1）消失点偏m差 （2）前景车辆depth偏差（3）检测性能偏差。这些说明所谓的“high-quality”数据误导感知模型。
% 揭露geometry 关系变化，像素上前景对齐，但是周围环境，也影响。    生成（很准的）前景inpaint到环境中。（不是车，是3d几何空间）
%如何构造准确的几何空间，
% mse loss： pixel-level align，忽略几何空间结构的保真和约束，因此几何空间错误
% ？
% 用RLHF框架引入几何关系约束。
% %现有RLHF 将样本看作整体，指标xx，不能直接用（需要pixel-align）的几何校正，所以设计RLSPF

%（1）geometry-perception feedback，精心设计reward function捕捉像素级别的矫正信号
%（2）（图像级别）Hierarchical Geometric Feedback：vanishing point， lane（车道线，地平线） ， 路面深度
%（3）（场景级别-occ）single-frame to video depth feedback： single frame， cross frame map 2.5% 

The rapid progress of autonomous driving (AD) systems~\cite{hu2023uniad,jiang2023vad,chen2024vadv2} has created a growing need for high-quality synthetic data. Recent diffusion-based video generation methods~\cite{wen2024panacea,yan2024drivingsphere,ma2024unleashing,gao2023magicdrive,gao2024magicdrivedit,jiang2025dive,jiang2025dive2} have achieved state-of-the-art visual realism, measured by metrics like FVD~\cite{unterthiner2019fvd}.
However, we identify a critical yet underexplored limitation: the generated videos often contain subtle yet impactful incorrect geometric relationships. This flaw not only misleads downstream perception and planning tasks but also undermines the reliability of models trained or evaluated using such data, significantly constraining their applicability in essential use cases such as simulation-based training~\cite{ma2024unleashing} and system validation~\cite{yan2024drivingsphere,yang2024drivearena}.

To investigate this geometric limitation, we conduct a series of targeted experiments. 
Evaluating 3D object detection on synthetic videos using BEVFusion~\cite{liang2022bevfusion} reveals a substantial performance drop compared to real data (mAP: 25.7 vs. 35.5). In contrast, 2D object detection using YOLOv5~\cite{jocher2020ultralytics} on the same data yields results comparable to real-world samples (mAP: 43.8 vs. 44.7).
These findings suggest that while current diffusion models~\cite{wen2024panacea,wang2024driving,jiang2025dive,gao2024magicdrivedit} preserve 2D appearance, indicating minimal image-level domain gap, yet they fail to capture accurate 3D scene structure. 
We attribute this primarily to underlying geometric inconsistencies.
To further verify this hypothesis and systematically quantify these distortions, we introduce GeoScores. 
This metric suite evaluates geometric fidelity by applying pre-trained perception models~\cite{depth_anything_v2,ravi2024sam2} to both synthetic videos and their corresponding real-world counterparts, using the outputs from real videos as reference ground truth.
Significant discrepancies between the two highlight geometric errors.
GeoScores (details are in \cref{sec:dataset}) reveals three major issues: (1) Vanishing point shifts, indicating incorrect global perspective; (2) Lane topology inconsistencies, reflecting misaligned road markings and implausible lane structures; and (3) Depth errors, particularly on road surfaces, signifying incorrect placement. (e.g., for a typical baseline, an average VP shift of 0.086 normalized units, a Lane F1-Score of only 0.792, and an average depth error of 1.822).
The significant deviations in these scores confirm that current "high-quality" synthetic data often suffers from pervasive geometric inaccuracies (\cref{fig:motivation}(a)).
% TODO 这三个指标什么意思 分别能隐含哪方面的信息
% TODO 把Score 列一下 得出个结论 生成图像的geo不对
% 从不同step的深度图可以看出来，geometry的生成是一个逐渐完善的过程，在早期step的时候（xx step之前），主要影响整体大结构geo的生成，在后期step的时候影响细节的geo的生成。所以整体end-to-end 的方式 无法对早期geo的生成做出准确的引导。 所以我们改成了采样window的方式训练，这样可以充分的对早期和后期进行引导。
% Despite their realistic appearance, these geometrically flawed videos mislead downstream perception models, particularly those relying on spatial reasoning.

Addressing these challenges, we present Reinforcement Learning with Geometric Feedback (RLGF), a novel framework that injects perception-model-driven geometric spatial constraints directly into the video generation process.
% Unlike conventional approaches~\cite{wen2024panacea} that rely on road sketches or 3D boxes to dictate generated content, RLGF leverages pre-trained AD perception models and designs a reward function to consistently improve the geometric integrity of videos generated by arbitrary video diffusion models~\cite{jiang2025dive,gao2024magicdrivedit} as a plug-and-play methodology.
%%%%%%%%%%%%
Unlike conventional approaches~\cite{wen2024panacea,wang2024driving,yan2024drivingsphere,jiang2025dive} that primarily rely on pixel-wise alignment, which often fail to explicitly enforce adherence to complex, underlying geometric principles, RLGF leverages dedicated, pre-trained AD perception models as reward providers to ensure geometric fidelity.

% Crucially, RLGF employs a novel, efficient, latent-space training strategy: Rewards are applied directly to noisy latent features within a sliding window of intermediate diffusion steps. This allows corrective feedback to be applied during the critical early-to-mid stages of denoising where core structures are formed, and significantly reduces the computational (GPU memory) burden compared to rewarding only final, decoded outputs. This enables RLGF to serve as a plug-and-play module that improves the geometric integrity of arbitrary video diffusion models~\cite{jiang2025dive,gao2024magicdrivedit}.
RLGF introduces two core technical innovations: Latent-Space Windowing Optimization and Hierarchical Geometric Alignment. 
% RLGF primarily comprises two key modules: Hierarchical Geometric Alignment (HGA) and Spatial-Temporal Occupancy Consensus.
% HGA achieves geometric consistency in autonomous driving scenes necessitates satisfying constraints across multiple scales, from local details like vanishing points to global structures such as lane topology. 
% Conventional methods~\cite{wen2024panacea} struggle to harmonize these multi-scale geometric relationships, often resulting in synthesized data with perspective distortions. 
% HGA addresses this by constructing a point-line-plane, multi-level feedback loop. It uses a perception model, operating on latent representations, to assess vanishing point, lane parsing, and depth estimation metrics. This reward-based feedback ensures that the multi-scale geometric structure within the synthesized data adheres to real-world physical principles.
% Spatial-Temporal Occupancy Consensus (STOC): The temporal coherence of autonomous driving videos relies not merely on appearance smoothness but critically on the plausible motion of dynamic objects within a 4D space-time volume. Existing frame-by-frame generation mechanisms inherently lack explicit modeling of scene-level motion dynamics. STOC leverages a pre-trained occupancy prediction model to impose 4D scene consistency constraints within the 3D spatial domain, significantly enhancing the compatibility of the synthesized data with downstream perception models.
Frist, we present Latent-Space Windowed Optimization. We observe that geometric structures in diffusion models emerge progressively across denoising steps: early steps (e.g., before step 10 in flow matching~\cite{lipman2022flow,liu2022rectified}) establish coarse global geometry, while later steps refine local details (\cref{fig:motivation}(c)). 
Training across the entire sampling process, as done in some prior RL-diffusion work~\cite{black2023vader} struggles to provide targeted guidance for these distinct phases. 
Therefore, we propose an efficient latent-space training strategy where rewards are applied directly to noisy latent features within a randomly sampled sliding window of intermediate diffusion steps. This approach significantly reduces computational (GPU memory) burden and, more importantly, allows for effective and targeted corrective feedback during both early (global structure formation) and late (detail refinement) stages of geometric synthesis.

%HGA aims to构建 point-line-plane, from feature-level to scene-level 的 multi-level feedback loop. For this 我们首先训练了一个多头多任务模型，named latent geometry perception model，来做vashing point detection，lane parsing和depth estimation。besides，我们也训练了一个latent occupancy prediction model 来预测scene-level voxels。这两个模型以diffusion process中的带噪latent feature和current timestamp作为输入，输出targe perception结果，因此我们也从训好的VAE encoder borrow 一些层introduce a micro-deocde module以增强对latent feature的特征捕捉能力。在latent space直接进行感知模型的预测大大节省了memory和计算开销。 After that, 为了充分挖掘perception models捕捉几何信息的能力，我们根据它们的输出设计多层级的reward，包括保持vashing point consistency，lane topology validity, and depth coherence的point-line-plane 的几何反馈和利用KL散度对语义特征对齐，在scene-level上做空间voxel的iou对齐。这样多角度的感知结果，多层级的reward设计保证了充分挖掘video的几何信息并合理feedback。

Secondly, RLGF features Hierarchical Geometric Reward (HGR), a multi-level feedback system designed to imbue generated videos with robust geometric fidelity and scene coherence.
HGA integrates signals from two specialized latent-space perception networks: a Latent Geometry Perception Model assessing vanishing point, lane, and depth cues, and a Latent Occupancy Prediction Model inferring 3D scene occupancy.
Both operate efficiently on noisy latents, aided by a lightweight micro-decode module to circumvent costly full decoding.
Leveraging these perception models, HGA then constructs its multi-level reward system.
For point-line-plane geometric feedback, we use the outputs of the perception model to: (1) enforce vanishing point consistency for accurate global perspective, (2) ensure lane topology validity for realistic road structure, and (3) promote depth coherence for correct surface and object geometry.
For scene-level occupancy feedback, outputs from the occupancy model are used to: (4) align intermediate semantic features using KL divergence for plausible scene evolution, and (5) maximize 3D occupancy IoU for accurate volumetric object layout and dynamics.
This structured approach, combining efficient latent-space perception with a multi-faceted reward design, ensures that comprehensive geometric and occupancy information is extracted and effectively fed back to guide the video diffusion model towards producing physically principled and geometrically sound autonomous driving scenarios.

Our contributions are fourfold:
\begin{itemize}

\item We are the first to systematically quantify the geometric distortion problem in autonomous driving video generation and propose the GeoScores metric for its evaluation.
\item We introduce RLGF, a novel paradigm that uses reinforcement learning with perception-based rewards applied efficiently within a sliding window in latent space, enabling plug-and-play geometric correction.
\item We design HGR which addresses geometric distortions by incorporating point-line-plane and scene-level occupancy multi-level geometric feedback derived from latent representations.
\item Extensive experiments on nuScenes demonstrate RLGF’s plug-and-play effectiveness across two baselines~\cite{jiang2025dive,gao2024magicdrivedit}, boosting 3D detection mAP by 12.7\% absolute while reduce geometry gap (via GeoScores) relative to real data. 
This work establishes a new paradigm for geometrically faithful synthetic data generation in autonomous driving systems.
\end{itemize}

\section{Related Work}
\subsection{Video Diffusion for Autonomous Driving}
The development of robust autonomous driving (AD) systems~\cite{hu2023uniad,jiang2023vad,chen2024vadv2} necessitates large volumes of diverse and realistic training data. 
% While real-world data is paramount, its collection and annotation are costly and time-consuming. 
% This has driven significant interest in synthetic data generation. Traditional simulators like CARLA [?] and LGSVL/Apollo Sim [?] provide data with high geometric and physical accuracy, derived from explicit 3D models and physics engines. However, they often struggle to match the visual fidelity and complexity of real-world scenes, leading to a significant domain gap when used for training perception models [?].
Learned generative model~\cite{kong20253d,yan2025olidm,goodfellow2020generative} have emerged as a powerful approach to synthesize such data by capturing the complex distributions of real-world driving scenarios.
Early efforts explored Generative Adversarial Networks (GANs)\cite{goodfellow2020generative,11119091} and Variational Autoencoders (VAEs)~\cite{kingma2013auto} for generating driving-related imagery.
More recently, diffusion models~\cite{ho2020denoising,nichol2021improved,rombach2022high,11186232,xin2025lumina,yi2024towards,yang2025matrix,dong2025panolora}, including latent diffusion models~\cite{rombach2022high}, have demonstrated state-of-the-art capabilities in generating high-fidelity images and videos~\cite{rombach2022high,blattmann2023align,blattmann2023svd,zheng2024opensora,wan2025,wang2024scantd,wang2025target,rongchao,ni2025recondreamer}. 
Based on this technique, generative models in autonomous driving have greatly advanced. For example, BEVGen~\cite{swerdlow2024bevgen} and BEVControl~\cite{yang2023bevcontrol} generate controllable street-view imagery from Bird's-Eye View (BEV) layouts or other structural conditions.
Further advancements, such as DriveDreamer~\cite{wang2024drivedreamer}, Magicdrive~\cite{gao2023magicdrive},DriveWM~\cite{wang2024driving}, Panacea~\cite{wen2024panacea}, DriveSphere~\cite{yan2024drivingsphere} and others~\cite{zhao2025drivedreamer2,ma2024unleashing,mei2024dreamforge,xie2024xdrive,jiang2025dive,li2024uniscene,ni2025wonderturbo,zeng2025FSDrive,zeng2024driving}, focus on generating coherent multi-camera driving videos, often conditioned on textual prompts, historical trajectories, or 3D assets. These models excel at producing visually compelling outputs that achieve high scores on perceptual metrics like Fréchet Video Distance (FVD)~\cite{unterthiner2019fvd}.

Despite achieving high visual realism, leading AD video generation models~\cite{wen2024panacea, gao2024magicdrivedit, jiang2025dive} frequently introduce subtle geometric and scene-level distortions (e.g., flawed perspectives, depth errors, unrealistic motion). These inconsistencies, often overlooked by human assessment, severely undermine performance on 3D perception tasks like object detection~\cite{liang2022bevfusion} and motion forecasting. RLGF addresses this critical gap by proposing a novel framework to instill multi-scale geometric and scene-level consistency within the generation process.

\subsection{Reinforcement Learning for Video Generation}
Fine-tuning generative models with reward signals, rooted in Reinforcement Learning from Human Feedback (RLHF) successes in LLMs (e.g., using PPO~\cite{schulman2017ppo} or DPO~\cite{rafailov2023dpo}), is increasingly applied to diffusion models. 
For image generation, methods \cite{black2023ddpo,wallace2024diffusiondpo,yang2024using,xu2023imagereward} align models with human preferences, and approaches~\cite{li2024controlnet++} use perception model feedback for content control.
This trend extends to video diffusion, where DPO-based techniques~\cite{yang2025ipo, zhang2024onlinevpo,yuan2024instructvideo} often utilize human preference data. While improving general video quality, these holistic preference scores typically lack the precise, local geometric feedback crucial for autonomous driving applications. Other methods~\cite{black2023vader,li2024t2vturbo,ni2025recondreamerrl,chen2025visrl} employ explicit reward models for video fine-tuning, offering more detailed signals. However, these rewards are generally not tailored to the specific multi-scale 3D geometric and physical plausibility demands of AD simulation, failing to adequately address distortions like incorrect perspective or depth error.

Our RLGF framework advances this by introducing explicit, quantifiable geometric and scene-level rewards from dedicated AD perception models. This provides targeted, locally-aware feedback to correct specific geometric inaccuracies, producing data suitable for rigorous AD tasks.

\section{Method}
\subsection{Preliminary: Limitations of Conditional Video Diffusion}
Current video diffusion models~\cite{blattmann2023svd,wen2024panacea,jiang2025dive} generate frames by gradually denoising latent variables through a Markov chain.  
Formally, given a condition $\bf{c}$ (e.g., road sketches or bounding boxes), the model learns to minimize the pixel-level reconstruction error during training:
\begin{equation}
    \mathcal{L}_\text{pixel}=\mathbb{E}_{x_0,\epsilon,t}[{||\epsilon-\epsilon_\theta(x_t,t,c)||}^2],
\end{equation}
where $x_t$ is the noisy sample at timestep $t$, $\epsilon$ is the ground truth noise, and $\epsilon_{\theta}$ is the denoising network. While effective for visual fidelity, this standard formulation inherently ignores crucial geometric-semantic correlations due to two limitations: (1) Pixel-wise Independence Assumption: The MSE loss treats each pixel as independent, failing to model higher-order geometric relationships (e.g., perspective consistency in 3D space). (2) Conditional Oversimplification: Existing methods concatenate geometric conditions $c$ with noisy latents, However, this primarily enforces local pixel alignment corresponding to the condition rather than guaranteeing the global geometric integrity or plausibility of the underlying 3D scene structure. These limitations motivate the need for explicit geometric guidance during generation.
\subsection{Reinforced Learning with Geometric Feedback (RLGF)}

Our primary objective is to enhance the geometric and spatio-temporal consistency of videos generated by conditional diffusion models, addressing the critical gap left by conventional training objectives that primarily focus on pixel-level visual fidelity. 

To achieve this, we introduce Reinforcement Learning with Geometric Feedback (RLGF). This framework refines a pre-trained video diffusion model, $\epsilon_\theta$, by guiding its generation process towards outputs that exhibit greater adherence to real-world geometric and physical principles. 
Unlike RLHF~\cite{ouyang2022rlhf,yang2024using}, which typically captures broad subjective preferences, RLGF is designed to incorporate specific, model-interpretable geometric constraints. 
The core idea is to leverage dedicated, pre-trained perception models to evaluate the generated videos $x_0$ and provide a reward signal that steers $\epsilon_\theta$ away from implausible outcomes.

Formally, given a well-trained video diffusion model $\epsilon_\theta$, the dataset $D_v$, the well-designed reward function $R(\cdot)$, we aim to maximize the objective:
\begin{equation}
    J(\theta)=\mathbb{E}_{c,v,x_0}[R(x_0,v)]
    \label{eq:rlgf_objective_x0_conceptual}
\end{equation}
where $c$ and $v$ denote the condition and real video from the dataset $D_v$, $x_0$ the the generated sample by $\epsilon_\theta(\cdot | c)$ given $c$, and the total reward $R(x_0,v)$ encapsulates assessments from our Hierarchical Geometric Alignment system, as detailed in Section~\ref{sec:hgr}.

\subsection{ Latent-Space Windowing RL Optimization}
Directly optimizing Equation~\ref{eq:rlgf_objective_x0_conceptual} by unrolling the entire 
$T$-step diffusion sampling process to obtain $x_0$ and then backpropagating the reward gradient $\nabla_{\theta}R(x_0,v)$ is computationally prohibitive due to large memory requirements and long sampling chains.

To efficiently apply RLGF, we propose Latent-Space Windowed RL Optimization. This strategy addresses the computational challenge of full rollouts and leverages insights into the progressive nature of structure formation in diffusion models.

\begin{figure}
    \centering
    \includegraphics[width=\linewidth]{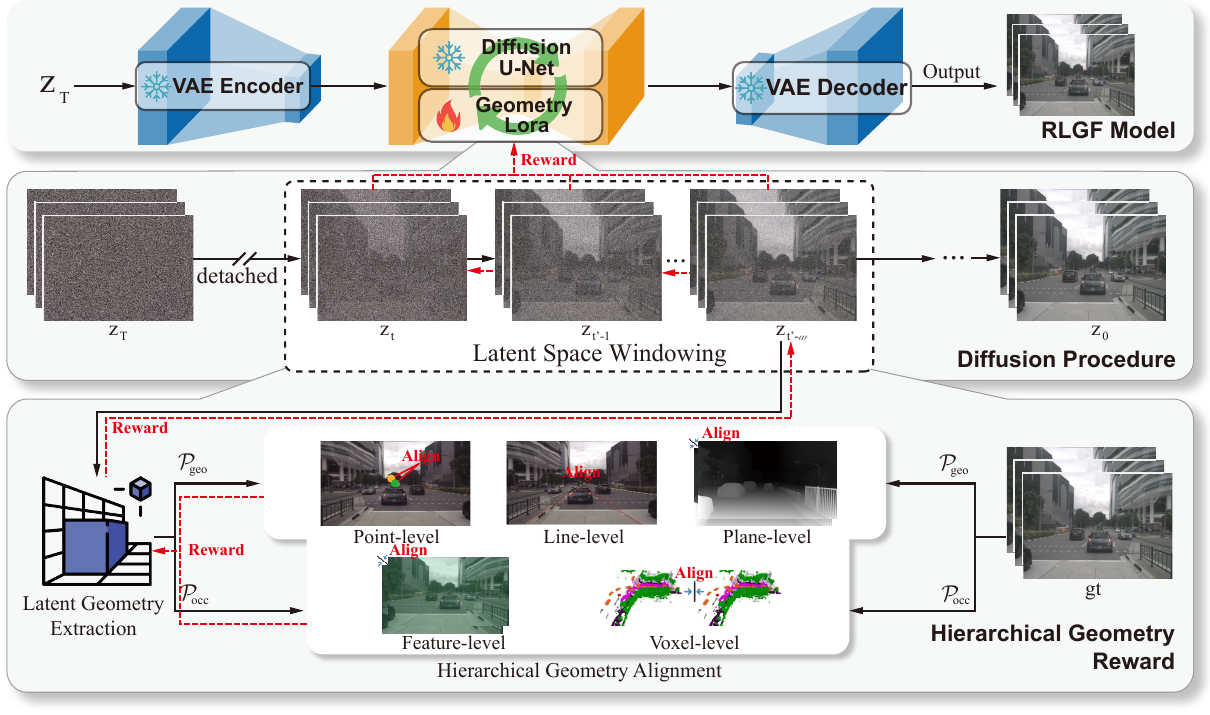}
    \vspace{-6mm}
    \caption{\textbf{Overview of the Reinforcement Learning with Geometric Feedback (RLGF) framework.} RLGF fine-tunes a frozen well-trained diffusion model via LoRA using rewards from a "Latent Space Windowing" scheme. Within this window, intermediate latents $z_{t'-w}$ are evaluated by frozen perception models $\mathcal{P}_{geo}$ (point-line-plane alignment) and $\mathcal{P}_{occ}$ (scene-level consistency) against a reference video. The resulting rewards ($R_{geo},R_{occ}$) generate gradients (red arrows) to update LoRA, improving geometric and temporal consistency. Black arrows: feed forward; dashed red: gradients.
}
    \vspace{-5mm}
    \label{fig:enter-label}
\end{figure}

Visualizations (e.g., \cref{fig:enter-label}(c) in Introduction) suggest that coarse global geometry is often established in earlier denoising steps (e.g. $t > T_{mid}$), while later steps refine details. Training across the entire sampling chain, as in some prior RL-diffusion work~\cite{black2023vader}, may not provide sufficiently targeted guidance for these distinct phases.

Instead of rewarding only the final output $x_0$, we provide feedback based on noisy latent features $z_t$ at intermediate denoising steps $t$, compared against reference latents $z_v$ derived from real videos.
Specifically, during the $T$-step sampling process $z_T \xrightarrow{} \dots \rightarrow z_0 $, we apply our reward functions $R$ within a sliding window of $w$ steps (e.g. from random start step $t'$ down to $k=t'-w$). Our perception models are designed to take $z_{k}$ and $k$ as input.

This approach is motivated by two factors: (1) Efficiency: It significantly reduces the computational graph for backpropagation. (2) Effectiveness: It allows targeted corrective signals at different stages of generation, crucial for both global structure and local detail.

The practical objective then becomes maximizing the expected reward obtained by evaluating the noisy latent $z_k$ (at step $k$ within the window) against the VAE-encoded latent $z_v$ paired with real video $v$:  
\begin{equation}
J_{practical}(\theta_{LoRA})=\mathbb{E}_{c,v , z_k} \left[  R(z_{k},z_v) \right]
\label{eq:rlgf_objective_practical}
\end{equation}
where and $z_v$ is the encoded latent feature using VAE ecnoder~\cite{rombach2022high}. The gradient with respect to the LoRA parameters $\theta_{LoRA}$ for a reward at step $k$ is then:

\begin{equation}
\nabla_{\theta_{LoRA}}R_k(z_k,z_v) = \frac{\partial R_k(z_k,z_v)}{\partial z_k} \cdot \frac{\partial z_k}{\partial \theta_{LoRA}}
\end{equation}
Throughout the RLGF fine-tuning, the perception models that constitute the reward function are pre-trained and their weights remain frozen. ensuring a consistent evaluation standard as $\epsilon_\theta $ via $ \theta_{LoRA}$ adapts to maximize geometric fidelity.

% Throughout the RLGF fine-tuning, the perception models that constitute the reward function are pre-trained and their weights remain frozen. This ensures a consistent evaluation standard as the diffusion model $\epsilon_\theta$ adapts to maximize geometric and spatio-temporal fidelity. 
% The specific designs of the $R_{geo}$ and $R_{occ}$ components are elaborated in the following subsections.
\subsection{Hierarchical Geometric Reward (HGR)}
\label{sec:hgr}
%消失点-车道线-地面和车辆面的深度度量能够反映图像的几何信息，我们需要端到端的感知模型来提取。previous方法没有将这些。我们创新型地将这几个子任务统一到一个端到端模型中，将这些任务都建模为dense prediction任务。
The Hierarchical Geometric Reward ($R$) is designed to  provide comprehensive, multi-level feedback on the geometric integrity and scene coherence of generated video latents. 
% Geometric consistency in autonomous driving scenes involves correct perspective (points/lines), accurate scene structure (planes/depth), and valid road layout (lanes). 
% To provide feedback at these different granularities, we leverage perception models capable of three key geometric tasks:
% (1) Vanishing Point Detection: Predicting the location of the scene's vanishing point.
% (2) Lane Parsing: Identifying and parameterizing lane lines.
% (3) Depth Estimation: Predicting the depth map for the scene.
%以实现点-线-面多粒度几何反馈
It evaluates consistency across point, line, plane, perceptual feature, and voxel-level representations, derived by applying specialized perception models to both the generated latent $z_k$ and the reference real-video latent $z_v$.

\subsubsection{Efficient Latent-Space Perception}
To avoid computationally expensive full VAE decoding at each step $k$ for perception, our perception models operate directly on latent features. Given a noisy video latent $z_k \in \mathbb{R}^{L \times C \times H' \times W'}$ (where $L$ is frames, $C$ channels, $H', W'$ latent dimensions) at diffusion step $k$, we first employ a lightweight Micro-Decode Module, $\mathcal{F}_{micro}$. This module, constructed using shallow layers from the VAE decoder, processes $z_k$ and $k$ to produce enhanced per-frame features:
\begin{equation}
    \mathbf{f}^f_k = \mathcal{F}_{micro}(z^f_k, k)
\end{equation} 
where $k$ is processed by the Fourier Embedding~\cite{zheng2024opensora,tancik2020fourier}.
$\mathbf{f}^f_k$ is suitable for downstream perception tasks, significantly reducing memory and computation. The same $\mathcal{F}_{micro}$ is applied to $z_v$ (with $k=0$) to obtain reference features $\mathbf{f}^f_v$.
\subsubsection{Perception Model Architectures}
We train two models that take these micro-decoded features $\{\mathbf{f}^f_k\}$ or $\{\mathbf{f}^f_v\}$ and timestep $k$ as input.

% Conventional approaches require computationally intensive VAE decoders to map diffusion latents $z_t \in \mathbb{R}^{H \times W \times C \times L}$ to pixel space for geometric perception, incurring substantial GPU memory overhead during fine-tuning.
% To address this, we design a Latent Geometry Perception Model, $\mathcal{P}_{geo}$ featuring the micro-decode module and unified multi-task architecture.

% We first borrow shallow layers from the decoder of pretrained VAEs to design the microdecode module $\mathcal{F}_{micro}$ for efficient feature disentanglement.
% Given the noisy latent feature $z_t$ at diffusion step $t$, we obtain:
% \begin{equation}
%     \bf{f}_t=\mathcal{F}_{micro}(z_t,t)
% \end{equation}
% where $\bf{f}_t$ denotes the decoded features.

\textbf{Latent Geometry Perception Model ($\mathcal{P}_{geo}$):}
This multi-task model processes per-frame features $\mathbf{f}^f$ to assess static 2.5D geometry. It uses a DINOv2~\cite{oquab2023dinov2} backbone followed by task-specific heads: 

(1) For \textbf{vanishing point detection}, we reformulate the task as heatmap regression to better capture positional uncertainty, following \cite{honda2021endvp}. The network head takes $\bf{g}$ as input and predicts a probability heatmap $H\in \mathbb{R}^{h \times w}$. The ground truth heatmap $H^{gt}$ is constructed as a 2D Gaussian centered at the annotated VP location, balancing localization precision and learning difficulty. 
We optimize using MSE:
\begin{equation}
    \mathcal{L}_{vp}=\|H-H^{gt}\|^2
\end{equation}
(2) The \textbf{lane parsing} branch complements this by detecting road markings through a topology-aware segmentation head, following \cite{ravi2024sam2}.

(3) As for the \textbf{depth estimation} task, we fine-tune the weight of \cite{depth_anything_v2} using a scale-invariant logarithmic (SiLog) loss:
\begin{equation}
    \mathcal{L}_{depth}=\frac{1}{n}\sum_id^2_i-\frac{\lambda}n^2({\sum_id_i})^2
\end{equation}
where $d_i=\text{log}y_i-\text{log}y_i^{gt}$, $y_i$ is the predicted results and $\lambda\in[0,1]$

Finally, we optimize $P_{geo}$ using the total loss $    \mathcal{L}_{geo}=\mathcal{L}_{vp}+\mathcal{L}_{lane}+\mathcal{L}_{depth}
$.

\textbf{Latent Occupancy Prediction Model ($\mathcal{P}_{occ}$):}
This model processes per-frame features $\mathbf{f}^f$ using an image backbone followed by a prediction head to infer the intermediate backbone features $\mathbf{feat}_{occ}$ and the 3D occupancy grid $O^i\in \mathbb{R}^{X \times Y \times Z}$ representing the occupied space around the ego vehicle for that frame. $\mathcal{P}_{occ}$ is trained based on~\cite{huang2024probabilistic}.

\subsubsection{Multi-Granularity Reward Signals}

The hierarchical geometric reward $R=R_{geo}+R_{occ}$ quantifies geometric fidelity and scene coherence through a multi-scale decomposition.

For $R_{geo}$, it addresses vanishing point consistency, lane topology validity, and depth coherence. Given the outputs $\left\{ p_{vp},L_{pred},D_{pred} \right\}$ from $\mathcal{P}_{geo}(z_k)$ and reference conditions $\left\{ v_{ref},L_{ref},D_{ref} \right\}$ from $\mathcal{P}_{geo}(z_v)$, we define:

\begin{equation}
    R_{geo}(z_k,z_v)=\underbrace{\lambda_{vp}r_{vp}(p_{vp},v_{ref})}_{\text{Point- level}}+\underbrace{\lambda_{lane}r_{lane}(p_{lane},L_{ref})}_{\text{Line- level}}+\underbrace{\lambda_{depth}r_{depth}(p_{depth},D_{ref})}_{\text{Plane-level}}
\end{equation}
where $p_{vp}$,$p_{vp}$,$p_{vp}$ are the predictions from $\mathcal{P}_{geo}$ for vanashing point, lanes and depth, respectively. $\lambda_{vp}$,$\lambda_{vp}$ and $\lambda_{vp}$ weights balancing the contribution of each task. 
The individual reward functions $r(\cdot)$ are designed to be high for geometrically accurate samples and low (or negative) for inconsistent ones: 
\begin{equation}
    r_{vp}(p_{vp},v_{ref})=-\|p_{vp}-v_{ref}\|_{2}^{2}
\end{equation}
where $v_{ref}$ is the ground-truth vanishing point in conditions.
\begin{equation}
    r_{lane}=\text{F1-Score}(L_{pred},L_{ref})
\end{equation}
where F1-Score indicates the F1-Scores.
% The Depth Reward $r_{depth}$ assesses the plausibility and consistency of the predicted depth map $p_{depth}$. 
\begin{equation}
    r_{depth}=-(\sqrt{(D_{pred } \odot M_{road} -D_{ref} \odot M_{road})^2}+\sqrt{(D_{pred } \odot M_{vehicle} -D_{ref} \odot M_{vehicle})^2} )
\end{equation}
% We specifically evaluate depth consistency for key scene elements: the road surface and vehicles. 
where pixel masks $M_{road}$ and $M_{vehicle}$ are generated from real data and identify road and vehicle regions, respectively.

After that, we define $R_{occ}(z_k,z_v)=r_{align}+r_{iou}$. $r_{align}$ encourages similarity in high-level scene interpretation by aligning distributions of intermediate occupancy features. For each frame $f$:
\begin{equation}
        r_{align_f} = -D_{KL}(p(\mathbf{feat}_{occ,f}^{real}) \| p(\mathbf{feat}_{occ,f}^{gen})),
    \end{equation}
    where $\mathbf{feat}_{occ,f}$ are backbone features from $\mathcal{P}_{occ}$. Distributions $p(\cdot)$ can be estimated (e.g., Gaussian fit over a batch).

$r_{occ}$ promotes accurate 3D structure and object layout. For each frame $f$:
    \begin{equation}
        r_{iou_f} = \text{IoU}(O_{f}^{gen}, O_{f}^{real}) = \frac{|O_{f}^{gen} \cap O_{f}^{real}|}{|O_{f}^{gen} \cup O_{f}^{real}|},
    \end{equation}
    where $O_f$ are the 3D occupancy grids from $\mathcal{P}_{occ}$.

\section{Experiments}
\subsection{Experimental Setup}
We conduct comprehensive experiments to validate the efficacy of our proposed Reinforcement Learning with Geometric Feedback (RLGF) framework. This section details the datasets, baseline models, evaluation metrics, and implementation specifics used in our evaluation.

\textbf{Datasets.} 
\label{sec:dataset}
Our experiments are primarily conducted on nuScenes~\cite{caesar2020nuscenes} using the official validation split. Its multi-camera setup and comprehensive annotations, including 3D object labels and HD maps, provide a challenging benchmark.
For this dataset, conditions $c$ for the diffusion models are extracted from ground truth annotations, simulating realistic inputs for controllable generation.
% \subsection{Datasets}

\textbf{Baselines.} We demonstrate the plug-and-play nature of RLGF by integrating it with two representative video diffusion models, referred to as MagicDrive-V2~\cite{gao2024magicdrivedit} and DiVE~\cite{jiang2025dive}. 
% These models cover a range of conditioning mechanisms and architectural designs prevalent in recent literature. 
% We use their publicly available pre-trained checkpoints as the starting point for our RLGF fine-tuning.

\textbf{Evaluation Metrics.} GeoScores: We use our proposed GeoScores suite:
Vanishing Point Error: NormDist between the predicted VP and the pseudo-ground truth VP derived from real data. Lower is better.
Lane Topology Score: F1-score for semantic segmentation of lane markings against ground truth lane masks. Higher is better.
Depth Error: We use RMSE between predicted depth for road surface regions and pseudo-ground truth depth. Lower is better.
% We also report an Aggregate GeoScore Gap (\%), representing the average percentage deviation of the synthetic data's GeoScores from those of real data, aiming for a smaller gap.
Downstream Task Performance:

3D Object Detection: We employ a strong BEV-based detector, BEVFusion~\cite{liang2022bevfusion}, trained solely on synthetic data generated by different methods or on real data. We report the standard nuScenes detection score (NDS) and mean Average Precision (mAP).

% 2D Object Detection: As a control, we use YOLOv5~\cite{jocher2020ultralytics} and report mAP, primarily to highlight the specific challenges in 3D geometry.
Visual Realism: We report Fréchet Video Distance (FVD)~\cite{unterthiner2019fvd} to ensure RLGF does not degrade the visual quality achieved by the baseline diffusion models.

\textbf{Implementation Details} are included in the supplemental materials.

\subsection{Performance of Pre-trained Perception Models}
% huge teacher model 's pseudo labels
We first verify the capabilities of our perception models ($\mathcal{P}_{geo}$ and $\mathcal{P}_{occ}$ which form the basis of our reward functions. 
These models operate directly on latent features $z_t$ and timestep $t$. Our goal here is to demonstrate their competence in extracting meaningful geometric and scene information from the latent domain.
  
 \begin{table*}[t]
  \centering
  \caption{\textbf{Performance of our Latent Geometry Perception Model ($\mathcal{P}_{geo}$) on the nuScenes validation split.} $\mathcal{P}_{geo}$ operates directly on micro-decoded latent features and is compared against representative pixel-space baselines.}
  \label{tab:p_geo_performance}
 \setlength{\tabcolsep}{18pt}
    \resizebox{\linewidth}{!}{
  \begin{tabular}{@{}llccc@{}}
    \toprule
    Task & Metric & Model / Method & Input Space & Performance \\
    \midrule
    \multirow{3}{*}{VP Detection} & \multirow{3}{*}{NormDist $\downarrow$} & URVP~\cite{liu2020unstructured} & Pixel & 0.045 \\
    & & VPD~\cite{honda2021end} & Pixel & 0.032 \\
    & & \textbf{$\mathcal{P}_{geo}$} & \textbf{Latent} & \textbf{0.024} \\
    \midrule
    \multirow{3}{*}{Lane Parsing} & \multirow{3}{*}{F1-Score $\uparrow$} & LaneATT~\cite{tabelini2021laneatt} & Pixel & 0.822 \\
    & & PriorLane~\cite{qiu2023priorlane} & Pixel & 0.879 \\
    & & \textbf{$\mathcal{P}_{geo}$} & \textbf{Latent} & 0.865 \\
    \midrule
    \multirow{2}{*}{Depth Estimation} & \multirow{2}{*}{RMSE $\downarrow$} & DepthAnything-v2~\cite{depth_anything_v2} & Pixel & \textbf{1.798} \\
    & & \textbf{$\mathcal{P}_{geo}$ } & \textbf{Latent} & 2.596 \\
    \bottomrule
  \end{tabular}}
\vspace{-2mm}
\end{table*}

 \begin{table*}[t]
  \caption{\textbf{Comparison with state-of-the-art video generation methods on the nuScenes validation set.} We evaluate visual quality (FVD), 3D Object Detection (3DOD) performance, and geometric fidelity (GeoScores).}
  \label{sample-table}
  \centering
   \setlength{\tabcolsep}{15pt}
    \resizebox{\linewidth}{!}{
  \begin{tabular}{c|c|cc|ccc}
    \toprule
    \multirow{2}{*}{Methods} & {Quality} & \multicolumn{2}{c|}{3DOD} & \multicolumn{3}{c}{GeoScore} \\
    \cline{2-7}
    &FVD& mAP& NDS & VP & Lane & Depth \\
    \hline
    Real Data &-& 35.53 & 41.20 & -  & - & - \\
    \hline
    Panacea~\cite{wen2024panacea} &139.0& 11.58 & 22.31 & -  & - & - \\
    Drive-WM~\cite{wen2024panacea} &122.7& 20.66 & - & -  & - & - \\
    MagicDrive-v2~\cite{jiang2025dive} &101.2& 18.95 & 21.10 & 0.092  & 0.787 & 1.732 \\
    DiVE~\cite{jiang2025dive} &68.4& 25.75 & 33.61 & 0.086  & 0.792 & 1.822 \\
    \hline
    MagicDrive-v2+Ours        &99.8& 23.21 & 27.80 & 0.079  & 0.854 & 0.983 \\
    DiVE+Ours                 &\textbf{67.6}& \textbf{31.42} & \textbf{36.07} & \textbf{0.068}  & \textbf{0.879} & \textbf{0.772} \\
    \bottomrule
  \end{tabular}}
    \vspace{-2mm}
\end{table*}

 \textbf{Latent Geometry Perception Model ($\mathcal{P}_{geo}$)} is a multi-task model responsible for assessing fine-grained geometric properties from latent features. Table~\ref{tab:p_geo_performance} presents its performance on the nuScenes validation split for vanishing point (VP) detection, lane parsing, and depth estimation, compared against established pixel-space methods. The results indicate that $\mathcal{P}_{geo}$ effectively captures these geometric cues from latent features. For instance, it achieves a normalized distance error of 0.024 for VP detection and an F1-score of 0.865 for lane parsing. 
Although its depth estimation RMSE (2.596) is higher than the specialized pixel space model DepthAnything-v2, it still provides a consistent measure of depth relationships.

 \textbf{Latent Occupancy Model ($\mathcal{P}_{occ}$)} is tasked with understanding 3D scene layout from sequences of latent features. Its performance on the Occ3D-nuScenes is shown in Table~\ref{tab:p_occ_performance}. 
$\mathcal{P}_{occ}$ achieves an overall mIoU of 29.96 when operating from latent representations. This level of performance, compared to a pixel-based method like FlashOcc (32.08 mIoU), demonstrates a solid capability to infer volumetric scene structure from the latent domain.

\subsection{Main Results: Improving Geometric Fidelity and Downstream Tasks} 
%geoscores 是用pretrained model在真实数据上的pseudo label来算，跟训练Pgeo的label一致。
Table~\ref{sample-table} shows RLGF's impact on the nuScenes validation set. RLGF consistently enhances geometric integrity (GeoScores) across baselines while maintaining or improving visual quality (FVD). For instance, DiVE + RLGF significantly improves all GeoScore components (VP error: 0.086 $\rightarrow$ 0.068; Lane F1: 0.792 
$\rightarrow$0.879; Depth RMSE: 1.822 
$\rightarrow$ 0.772). 

\begin{figure}[t]
     \centering
     \includegraphics[width=0.9\linewidth]{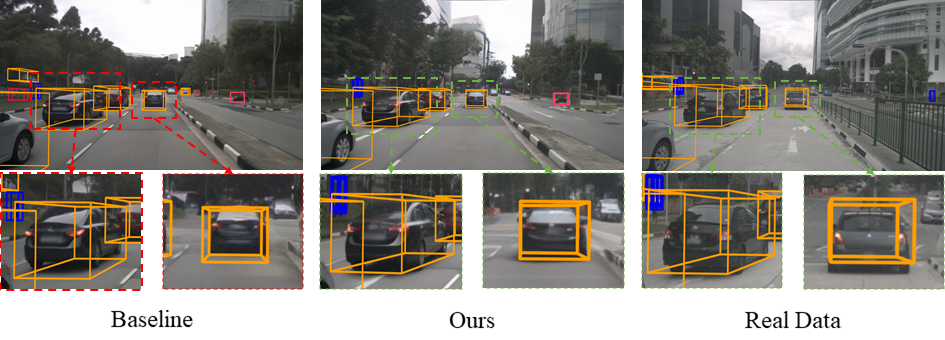}
     \vspace{-3mm}
     \caption{\textbf{Qualitative comparison of 3D object bounding box alignment.} RLGF-enhanced video exhibits much-improved 3D box alignment, closely matching the geometry implied by the scene.}
     \label{fig:vis}
         \vspace{-2mm}
 \end{figure}

Crucially, this geometric enhancement translates to substantial 3DOD gains: DiVE + RLGF boosts mAP from 25.75 to 31.42 and NDS from 33.61 to 36.07, markedly closing the gap to real data performance. Similar improvements are seen for MagicDrive-v2 + RLGF (mAP: 18.95 
$\rightarrow$
 23.21). These results underscore RLGF's effectiveness as a plug-and-play module for improving both geometric soundness and downstream utility of synthetic videos. 
 We also present the qualitative result in \cref{fig:vis} about the detection results on our synthetic data and baseline~\cite{jiang2025dive} and the real data.
 (More qualitative visualization results and further analysis are included in the supplementary materials).

\subsection{Ablation Study}
We conduct ablation studies on nuScenes using DiVE as the baseline to understand the contributions of different HGA reward components. Results are shown in Table~\ref{tab:ablation_hga}.
 Incrementally adding reward components ($r_{vp}$,$r_{lane}$,$r_{depth}$, and occupancy rewards ($r_{align}+r_{iou}$) progressively improves 3DOD performance (Table~\ref{tab:ablation_hga}). For example, adding point-line-plane rewards (ID: 3) boosts DiVE's mAP to 27.12. Incorporating only occupancy rewards (ID: 4) yields 28.06 mAP. The full HGA system, combining all five components, achieves the highest mAP (31.42) and NDS (36.07), highlighting the synergistic benefits of our comprehensive multi-level feedback. (More qualitative visualization results and further analysis are included in the supplementary materials).

\begin{table*}[t] % Use table* to span full text width for the container of minipages
  \begin{minipage}[t]{0.40\textwidth} % Adjust width as needed (e.g., 0.48\textwidth)
    \centering
    % \captionsetup{width=0.9\linewidth} % Optional: makes caption width slightly less than minipage
    \caption{\textbf{Performance of our Latent Occupancy Prediction Model ($\mathcal{P}_{occ}$) on the Occ3D-nuScenes.} $\mathcal{P}_{occ}$ predicts 3D occupancy grids from sequences of micro-decoded latent features. \textbf{Vehicle} and \textbf{Dri. Sur} indicate IoU of vehicles and driving surface.}
    \label{tab:p_occ_performance}
    % \vspace{-2mm} % Optional: reduce space between caption and table
        \resizebox{\linewidth}{!}{
    \begin{tabular}{@{}lccc@{}}
      \toprule
      Method &  Vehicle & Dri. Sur&  mIoU $\uparrow$  \\
      \midrule
      FlashOcc~\cite{yu2023flashocc} & 43.2& 72.2 & 32.08  \\
      \midrule
      \textbf{$\mathcal{P}_{occ}$} & 37.9 &65.7 & 29.96 \\
      \bottomrule
    \end{tabular}
    }
  \end{minipage} % \hfill will push the minipages apart to fill the line
  \hspace{2mm}
  \begin{minipage}[t]{0.59\textwidth} % Adjust width as needed
    \centering
    % \captionsetup{width=0.9\linewidth} % Optional
    \caption{\textbf{Ablation study of HGA reward components within RLGF on nuScenes.}} % Simplified caption for space
    \label{tab:ablation_hga}
    % \vspace{-2mm} % Optional
        \resizebox{\linewidth}{!}{
    \begin{tabular}{@{}c|ccccc|cc@{}}
      \toprule
      \multirow{2}{*}{\textbf{ID}} & \multicolumn{5}{c|}{HGA Rewards} & \multicolumn{2}{c}{3DOD} \\
      \cmidrule(lr){2-6} \cmidrule(lr){7-8}
      & $r_{vp}$ & $r_{lane}$ & $r_{depth}$ & $r_{align}$ & $r_{iou}$ & mAP$\uparrow$ & NDS$\uparrow$ \\ % Abbreviated reward names
      \midrule
      DiVE~\cite{jiang2025dive} &            &            &             &             &           & 25.75 & 33.61 \\
      \midrule
      1 & \checkmark &            &             &             &           & 26.31 & 33.66 \\
      2 & \checkmark & \checkmark &             &             &           & 26.93 & 33.98\\
      3 & \checkmark & \checkmark & \checkmark  &             &           & 27.12 & 34.82 \\
      4 &            &            &             & \checkmark  & \checkmark & 28.06 & 35.11 \\
      % Add more rows for full HGA and removing components if you have them
      \midrule
      Full& \checkmark & \checkmark & \checkmark  & \checkmark  & \checkmark& $\textbf{31.42}$ & $\textbf{36.07}$ \\
      \bottomrule
    \end{tabular}
    }
  \end{minipage}
\end{table*}

\section{Conclusion}
In this work, we addressed the critical issue of geometric distortions in diffusion-based video generation for autonomous driving. We introduced GeoScores for quantitative evaluation and proposed Reinforcement Learning with Geometric Feedback (RLGF), a novel framework to enhance the geometric integrity of synthetic videos. 
RLGF, through its Hierarchical Geometric Alignment (HGA) module which incorporates multi-level geometric and scene occupancy feedback, effectively injects perception-driven constraints into the generation process by fine-tuning pre-trained diffusion models. Our experiments demonstrate that RLGF significantly improves geometric fidelity across multiple baselines and, crucially, boosts downstream 3D object detection performance by up to , substantially closing the gap with real-data performance. This work establishes a new direction for generating more reliable and task-aware synthetic data for autonomous systems.

%%%%%%%%%%%%%%%%%%%%%%%%%%%%%%%%%%%%%%%%%%%%%%%%%%%%%%%%%%%%

\appendix
This supplementary material provides additional details to support the findings presented in our main paper. We include: (1) comprehensive implementation specifics for our RLGF framework and the pre-trained perception models; further details on the GeoScores metric computation; (2) additional experimental results (3) a discussion on the limitations of our current work and potential future directions.

\section{Detailed Implementation Details}
This section elaborates on the implementation details of our proposed RLGF framework, the pre-trained perception models ($\mathcal{P}_{geo} $ and $\mathcal{P}_{occ}$) , and the experimental setup.
\subsection{Dataset Preparation}
\label{sec:dataset}
All experiments, including the pre-training of our perception models ($\mathcal{P}_{geo} $ and $\mathcal{P}_{occ}$) and the fine-tuning of diffusion models with RLGF, are conducted using the nuScenes dataset~\cite{caesar2020nuscenes}. We primarily utilize the official training and validation splits. While nuScenes provides rich annotations like 3D bounding boxes and HD maps, it does not directly offer ground truth labels for vanishing points (VP), dense segmentation masks for all relevant classes (like fine-grained lanes beyond HD map polylines), or per-pixel depth maps required by our $\mathcal{P}_{geo}$. Therefore, we generate high-quality pseudo-labels for these tasks using strong, pre-existing perception models, as detailed below. 
These pseudo-labels serve as the training targets for our latent-space perception models.

\textbf{Depth Pseudo-Labels:} To obtain dense depth information for training the depth estimation component of $\mathcal{P}_{geo}$, we utilize Depth Anything V2 (vit-l version)~\cite{depth_anything_v2}. This state-of-the-art monocular depth estimation model is applied to all images in the nuScenes training set to generate per-pixel depth maps. These output depth maps serve as the pseudo-ground truth for our latent depth estimation task.

\textbf{Semantic Segmentation Pseudo-Labels (Lanes, Road Surface, Vehicles):}
For precise segmentation masks of various scene elements, we employ Grounded-SAM-2\cite{ren2024grounded,ravi2024sam2}. For the lanes, the model is prompted to accurately segment visible lane markings. The resulting binary segmentation masks are used as pseudo-ground truth for training the lane parsing head of $\mathcal{P}_{geo}$.
For the road surface and vehicle masks, SAM-2 is also utilized to generate segmentation masks for road surfaces and vehicles.

\textbf{Vanishing Point Pseudo-Labels from Lane Masks, following~\cite{honda2021endvp}:} With accurate lane segmentation masks obtained via SAM-2 (as described above), we derive vanishing point pseudo-labels through a geometric procedure. For each detected lane marking in a frame:
The center point of the lane marking is calculated from its left and right edges (derived from the SAM-2 segmentation mask) for every horizontal image line at 5-pixel intervals.
These extracted center points are grouped to represent the centerline of each individual lane marking.
Robust curve fitting (e.g., RANSAC with a line model) is applied to these centerlines.
The intersection point of multiple fitted lane centerlines is then computed to determine the scene's vanishing point. This computed VP serves as the pseudo-ground truth for the VP detection task.

The use of these high-quality pseudo-labels enables us to train effective latent-space perception models tailored to the nuScenes domain, which subsequently provide the nuanced reward signals for our RLGF framework. The conditions $c$ for the main diffusion models (e.g., semantic 3D boxes for some baselines) are derived from the original nuScenes ground truth annotations.

\subsection{Perception Model Architectures and Pre-training}
\textbf{Micro-Decode Module($\mathcal{F}_{micro}$)}: The $\mathcal{F}_{micro}$ module is constructed using the first upper block of the official VAE decoder from the OpenSora~\cite{zheng2024opensora} used by our baseline video diffusion models.
The input of $\mathcal{F}_{micro}$ is the noisy latent feature $z_k^f$ for a frame $f$ with the timestep $k$. The same $\mathcal{F}_{micro}$ architecture is used when processing the reference real-video latent $z_v$ (with $k$ typically set to 0).

\textbf{Latent Geometry Perception Model($\mathcal{P}_{geo}$)}: We use a pre-trained DINOv2-ViT-S/14~\cite{oquab2023dinov2} as the backbone feature extractor and the pre-trained weight from DepthAnything-V2~\cite{depth_anything_v2}.
$\mathcal{P}_{geo}$ is trained for 50 epochs on the nuScenes training split using the AdamW optimizer with a learning rate of $5\times10^{-5}$ and a batchsize of 16. We use 8$\times$ NVIDIA A100 GPUs to cover the experiment. 

\textbf{Latent Occupancy Prediction Model($\mathcal{P}_{occ}$)}:
$\mathcal{P}_{occ}$ is trained on occ3D-nuscenes dataset for 24 epochs using AdamW optimizer following \cite{yu2023flashocc,huang2024probabilistic}.
\subsection{RLGF Fine-tuning Details}
Baselines: We used publicly available checkpoints for MagicDrive-V2~\cite{gao2024magicdrivedit} and DiVE~\cite{jiang2025dive}.

LoRA Configuration: For LoRA, we applied it to the attention layers (Q, K, V projections) of the DiT backbone in the diffusion models. 
We used a rank $r=16$ following ~\cite{black2023vader}.

Latent-Space Windowed Optimization: The window size $w$ is set to 5.
The starting step $t'$ for the window was randomly sampled from the range $[8,30]$, with $T=30$ is the total number of diffusion steps.

Reward Weights:
We set $\lambda_{vp}=0.1,\lambda_{lane}=0.1,\lambda_{depth}=0.5$. These weights were determined empirically based on early experiments on a small validation subset, aiming to balance the scale of individual reward components and their perceived impact on generation quality.

We use AdamW with a learning rate of $1\times 10^{-4}$ and a batchsize of 1 with 8 frames per video clip.

\subsection{GeoScores Metric Details}

This section provides further clarification on the computation of our GeoScores components. For all GeoScores, the "reference ground truth" is derived by applying the corresponding pre-trained perception model to the \textit{real} video data, following~\cref{sec:dataset}.
 The score then measures the deviation of the synthetic video's perception output from this real-data-derived reference.

\textbf{Vanishing Point Error (VP$\downarrow$):}
Calculated as the L2 Normalized Distance (NormDist) between the calculated VP on a synthetic frame and the VP calculated on a real frame.
\textbf{Lane Topology Score (Lane$\uparrow$):}
Calculated as the F1 score for semantic segmentation of lane markings. The predictions is from Grounded-SAM2~\cite{ren2024grounded,ravi2024sam2} on the synthetic frame, and the target is applied to the real frame.
\textbf{Depth Error (Depth)$\downarrow$:} Calculated as the Root Mean Squared Error (RMSE)  between the depth map predicted by Depth Anything V2~\cite{depth_anything_v2} for road surface regions on a synthetic frame and the depth map for the same regions on the real frame. Road surface masks are obtained from SAM-2~\cite{ravi2024sam2}.

\section{Additional Experiment Results}
% Technical appendices with additional results, figures, graphs and proofs may be submitted with the paper submission before the full submission deadline (see above), or as a separate PDF in the ZIP file below before the supplementary material deadline. There is no page limit for the technical appendices.
\subsection{2D Object Detection Results}
%额外的检测器
\begin{figure}
    \centering
    \includegraphics[width=\textwidth]{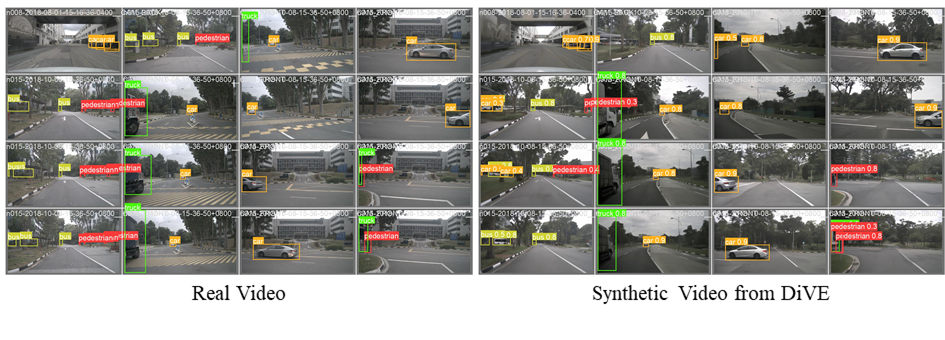}
    \vspace{-10mm}
    \caption{\textbf{Left:} Detection results on a real nuScenes image. \textbf{Right:} Detection results on a corresponding synthetic image generated by the DiVE baseline. Bounding boxes indicate detected objects (primarily vehicles).}
    \label{fig:2d_detection_comparison}
\end{figure}
To illustrate that current diffusion models like DiVE can generate visually realistic data with minimal 2D domain gap for certain tasks, we present qualitative 2D object detection results. Figure~\ref{fig:2d_detection_comparison} shows outputs from a YOLOv5~\cite{jocher2020ultralytics} detector applied to (a) real nuScenes data and (b) synthetic data generated by the DiVE baseline. The detector is pre-trained on a large-scale dataset (e.g., COCO) and then fine-tuned on real nuScenes training data.

As observed in Figure~\ref{fig:2d_detection_comparison}, the 2D detection performance on DiVE-generated synthetic data is qualitatively very similar to that on real data. Objects are generally detected with comparable confidence and bounding box accuracy. This visual consistency aligns with our quantitative findings (mAP: 43.8 on synthetic vs. 44.7 on real, as mentioned in the Introduction), suggesting that the semantic content and 2D appearance features necessary for 2D detection are well-preserved in the synthetic videos. This further reinforces our hypothesis that the primary limitation of such synthetic data lies in its 3D geometric fidelity, which is specifically addressed by our RLGF framework.

\subsection{Extended 3D Object Detection Results on Multiple Detecters}
To further demonstrate the generalizability of the improvements conferred by RLGF, we evaluated the generated synthetic data using an additional state-of-the-art 3D object detector, StreamPETR~\cite{wang2023exploring}, alongside the BEVFusion results presented in the main paper. Table~\ref{tab:supplementary_3dod_details} presents the performance (mAP and NDS on nuScenes validation) for StreamPETR and the average performance across both BEVFusion and StreamPETR. Both detectors were trained from scratch solely on the respective synthetic data or real data.

\begin{table}[t]
\centering
\caption{Detailed 3D Object Detection (3DOD) performance on nuScenes validation using StreamPETR~\cite{wang2023exploring}. RLGF is applied to MagicDrive-v2 and DiVE.}
\label{tab:supplementary_3dod_details}
\setlength{\tabcolsep}{5pt} % Adjust as needed
% \resizebox{\textwidth}{!}{ % May be needed if too wide
  \begin{tabular}{c|c|cc|cc}
    \toprule
    \multirow{2}{*}{Methods} & {Quality} & \multicolumn{2}{c|}{BevFusion} &\multicolumn{2}{c}{StreamPETR} \\
    \cline{2-6}
    &FVD& mAP& NDS & mAP& NDS  \\
    \hline
    Real Data &-& 35.53 & 41.20& 38.01 & 49.02  \\
    \hline
    Panacea~\cite{wen2024panacea} &139.0& 11.58 & 22.31 & - & - \\
    Drive-WM~\cite{wen2024panacea} &122.7& 20.66 & -& - & -  \\
    MagicDrive-v2~\cite{jiang2025dive} &101.2& 18.95 & 21.10& 22.77 & 28.93  \\
    DiVE~\cite{jiang2025dive} &68.4& 25.75 & 33.61& 29.19 & 36.23  \\
    \hline
    MagicDrive-v2+Ours        &99.8& 23.21 & 27.80& 26.01 & 35.64  \\
    DiVE+Ours                 &\textbf{67.6}& \textbf{31.42} & \textbf{36.07} & \textbf{33.94} & \textbf{39.68} \\
    \bottomrule
  \end{tabular}
% } % End resizebox
\end{table}

\subsection{ Robustness to Different Downstream Detectors:} To demonstrate generality, we evaluated our method on the stronger \textbf{StreamPETR} detector. RLGF achieved a substantial +4.75\% mAP gain, proving our geometric improvements are robust and benefit diverse downstream architectures.

\begin{table}[t]
    \centering
    \begin{tabular}{c|c}
    \toprule
          \textbf{Method} & \textbf{mAP (BEVFusion)}\\
          \midrule
          
         DiVE [16] (Baseline) & 25.75 \\
+ Detector Reward & 26.51 (+0.76) \\
\textbf{+ Ours (RLGF)} & \textbf{31.42 (+5.67)} \\
\bottomrule
    \end{tabular}
    \caption{Comparison of reward signal effectiveness on the DiVE baseline. Our Hierarchical Geometric Reward (HGR) is significantly more effective than a high-level detector-based reward.}
    \label{tab:detec}
\end{table}

\subsection{Ablation Study about Hyerparameters.} 
We provide the detailed ablation results here for clarity, including the RL window size, reward weights, and the starting step of the sliding window. All ablations are performed on the DiVE baseline and evaluated with BEVFusion.

\begin{table}[t]
\centering
\caption{Ablation studies on key RLGF hyperparameters.}
\label{tab:hyperparam_ablation}
\begin{tabular}{lc|c}
\toprule
\textbf{Hyperparameter} & \textbf{Value} & \textbf{3D Detection mAP} \\
\midrule
\multirow{3}{*}{\textbf{Window Size (w)}} & 3 & 30.89 \\
& \textbf{5 (Ours)} & \textbf{31.42} \\
& 8 & 31.25 \\
\hline
\multirow{2}{*}{\textbf{Reward Weights ($\lambda$)}}
& Equal Weights (all 0.2) & 30.76 \\
& \textbf{Balanced (Ours)} & \textbf{31.42} \\
\hline
\multirow{3}{*}{\textbf{Window Range Position (
$t'$)}} & Early (Random in [20, 30]) & 30.55 \\
& \textbf{Mid (Random in [8, 30]) (Ours)} & \textbf{31.42} \\
& Late (Random in [1, 15]) & 29.91 \\
\bottomrule
\end{tabular}
\end{table}

\section{Limitations and Future Work}
This section discusses the current limitations of our RLGF framework and GeoScores metric, alongside potential avenues for future research.

\textbf{Dependence on Perception Models:} RLGF's performance is inherently tied to the accuracy and robustness of the pre-trained perception models ($\mathcal{P}_{geo},\mathcal{P}_{occ}$). Biases or errors in these models could propagate into the reward signal and mislead the generation process. Future work could explore jointly training or adapting perception models during RLGF, or using ensembles.

\textbf{Computational Cost:} While Latent-Space Windowed Optimization significantly reduces costs compared to full rollouts, RL-based fine-tuning remains more computationally intensive than standard diffusion model training. Exploring more sample-efficient RL algorithms or distillation techniques could be beneficial.

\textbf{Reward Design and Balancing:} The current HGA reward combines five components with manually tuned weights. Optimizing these weights automatically or learning a more adaptive reward function is a promising direction. Furthermore, incorporating even more diverse geometric or physical constraints (e.g., collision avoidance, traffic rule adherence) could further enhance realism.

\textbf{Generalization:} While demonstrated on nuScenes, further investigation is needed to assess RLGF's generalization capabilities across diverse datasets, environmental conditions (e.g., adverse weather, night scenes not well-represented in training), and different diffusion model architectures.

\textbf{GeoScores Scope:} Current GeoScores focus on camera-based geometric aspects. Expanding them to include LiDAR consistency or multi-modal geometric agreement could provide a more holistic evaluation.

%%%%%%%%%%%%%%%%%%%%%%%%%%%%%%%%%%%%%%%%%%%%%%%%%%%%%%%%%%%%
\bibliographystyle{plain}
\bibliography{ref}

\begin{thebibliography}{10}

\bibitem{black2023vader}
Alexander Black, Simon Jenni, Tu~Bui, Md~Mehrab Tanjim, Stefano Petrangeli, Ritwik Sinha, Viswanathan Swaminathan, and John Collomosse.
\newblock Vader: video alignment differencing and retrieval.
\newblock In {\em Proceedings of the IEEE/CVF International Conference on Computer Vision}, pages 22357--22367, 2023.

\bibitem{black2023ddpo}
Kevin Black, Michael Janner, Yilun Du, Ilya Kostrikov, and Sergey Levine.
\newblock Training diffusion models with reinforcement learning.
\newblock {\em arXiv preprint arXiv:2305.13301}, 2023.

\bibitem{blattmann2023svd}
Andreas Blattmann, Tim Dockhorn, Sumith Kulal, Daniel Mendelevitch, Maciej Kilian, Dominik Lorenz, Yam Levi, Zion English, Vikram Voleti, Adam Letts, et~al.
\newblock Stable video diffusion: Scaling latent video diffusion models to large datasets.
\newblock {\em arXiv preprint arXiv:2311.15127}, 2023.

\bibitem{blattmann2023align}
Andreas Blattmann, Robin Rombach, Huan Ling, Tim Dockhorn, Seung~Wook Kim, Sanja Fidler, and Karsten Kreis.
\newblock Align your latents: High-resolution video synthesis with latent diffusion models.
\newblock In {\em Proceedings of the IEEE/CVF conference on computer vision and pattern recognition}, pages 22563--22575, 2023.

\bibitem{caesar2020nuscenes}
Holger Caesar, Varun Bankiti, Alex~H Lang, Sourabh Vora, Venice~Erin Liong, Qiang Xu, Anush Krishnan, Yu~Pan, Giancarlo Baldan, and Oscar Beijbom.
\newblock nuscenes: A multimodal dataset for autonomous driving.
\newblock In {\em Proceedings of the IEEE/CVF conference on computer vision and pattern recognition}, pages 11621--11631, 2020.

\bibitem{chen2024vadv2}
Shaoyu Chen, Bo~Jiang, Hao Gao, Bencheng Liao, Qing Xu, Qian Zhang, Chang Huang, Wenyu Liu, and Xinggang Wang.
\newblock Vadv2: End-to-end vectorized autonomous driving via probabilistic planning.
\newblock {\em arXiv preprint arXiv:2402.13243}, 2024.

\bibitem{chen2025visrl}
Zhangquan Chen, Xufang Luo, and Dongsheng Li.
\newblock Visrl: Intention-driven visual perception via reinforced reasoning.
\newblock {\em arXiv preprint arXiv:2503.07523}, 2025.

\bibitem{dong2025panolora}
Zeyu Dong, Yuyang Yin, Yuqi Li, Eric Li, Hao-Xiang Guo, and Yikai Wang.
\newblock Panolora: Bridging perspective and panoramic video generation with lora adaptation.
\newblock {\em arXiv preprint arXiv:2509.11092}, 2025.

\bibitem{gao2024magicdrivedit}
Ruiyuan Gao, Kai Chen, Bo~Xiao, Lanqing Hong, Zhenguo Li, and Qiang Xu.
\newblock Magicdrivedit: High-resolution long video generation for autonomous driving with adaptive control.
\newblock {\em arXiv preprint arXiv:2411.13807}, 2024.

\bibitem{gao2023magicdrive}
Ruiyuan Gao, Kai Chen, Enze Xie, Lanqing Hong, Zhenguo Li, Dit-Yan Yeung, and Qiang Xu.
\newblock Magicdrive: Street view generation with diverse 3d geometry control.
\newblock {\em arXiv preprint arXiv:2310.02601}, 2023.

\bibitem{goodfellow2020generative}
Ian Goodfellow, Jean Pouget-Abadie, Mehdi Mirza, Bing Xu, David Warde-Farley, Sherjil Ozair, Aaron Courville, and Yoshua Bengio.
\newblock Generative adversarial networks.
\newblock {\em Communications of the ACM}, 63(11):139--144, 2020.

\bibitem{ho2020denoising}
Jonathan Ho, Ajay Jain, and Pieter Abbeel.
\newblock Denoising diffusion probabilistic models.
\newblock {\em Advances in neural information processing systems}, 33:6840--6851, 2020.

\bibitem{honda2021endvp}
Hiroto Honda, Motoki Kimura, Takumi Karasawa, and Yusuke Uchida.
\newblock End-to-end monocular vanishing point detection exploiting lane annotations.
\newblock {\em arXiv preprint arXiv:2108.13699}, 2021.

\bibitem{honda2021end}
Hiroto Honda, Motoki Kimura, Takumi Karasawa, and Yusuke Uchida.
\newblock End-to-end monocular vanishing point detection exploiting lane annotations.
\newblock {\em arXiv preprint arXiv:2108.13699}, 2021.

\bibitem{hu2023uniad}
Yihan Hu, Jiazhi Yang, Li~Chen, Keyu Li, Chonghao Sima, Xizhou Zhu, Siqi Chai, Senyao Du, Tianwei Lin, Wenhai Wang, et~al.
\newblock Planning-oriented autonomous driving.
\newblock In {\em Proceedings of the IEEE/CVF conference on computer vision and pattern recognition}, pages 17853--17862, 2023.

\bibitem{huang2024probabilistic}
Yuanhui Huang, Amonnut Thammatadatrakoon, Wenzhao Zheng, Yunpeng Zhang, Dalong Du, and Jiwen Lu.
\newblock Probabilistic gaussian superposition for efficient 3d occupancy prediction.
\newblock {\em arXiv preprint arXiv:2412.04384}, 2024.

\bibitem{jiang2023vad}
Bo~Jiang, Shaoyu Chen, Qing Xu, Bencheng Liao, Jiajie Chen, Helong Zhou, Qian Zhang, Wenyu Liu, Chang Huang, and Xinggang Wang.
\newblock Vad: Vectorized scene representation for efficient autonomous driving.
\newblock In {\em Proceedings of the IEEE/CVF International Conference on Computer Vision}, pages 8340--8350, 2023.

\bibitem{jiang2025dive}
Junpeng Jiang, Gangyi Hong, Miao Zhang, Hengtong Hu, Kun Zhan, Rui Shao, and Liqiang Nie.
\newblock Dive: Efficient multi-view driving scenes generation based on video diffusion transformer.
\newblock {\em arXiv preprint arXiv:2504.19614}, 2025.

\bibitem{jiang2025dive2}
Junpeng Jiang, Gangyi Hong, Miao Zhang, Hengtong Hu, Kun Zhan, Rui Shao, and Liqiang Nie.
\newblock Dive: Efficient multi-view driving scenes generation based on video diffusion transformer.
\newblock {\em arXiv preprint arXiv:2504.19614}, 2025.

\bibitem{jocher2020ultralytics}
Glenn Jocher, Alex Stoken, Jirka Borovec, Liu Changyu, Adam Hogan, Laurentiu Diaconu, Jake Poznanski, Lijun Yu, Prashant Rai, Russ Ferriday, et~al.
\newblock ultralytics/yolov5: v3. 0.
\newblock {\em Zenodo}, 2020.

\bibitem{kingma2013auto}
Diederik~P Kingma, Max Welling, et~al.
\newblock Auto-encoding variational bayes, 2013.

\bibitem{kong20253d}
Lingdong Kong, Wesley Yang, Jianbiao Mei, Youquan Liu, Ao~Liang, Dekai Zhu, Dongyue Lu, Wei Yin, Xiaotao Hu, Mingkai Jia, et~al.
\newblock 3d and 4d world modeling: A survey.
\newblock {\em arXiv preprint arXiv:2509.07996}, 2025.

\bibitem{li2024uniscene}
Bohan Li, Jiazhe Guo, Hongsi Liu, Yingshuang Zou, Yikang Ding, Xiwu Chen, Hu~Zhu, Feiyang Tan, Chi Zhang, Tiancai Wang, et~al.
\newblock Uniscene: Unified occupancy-centric driving scene generation.
\newblock {\em arXiv preprint arXiv:2412.05435}, 2024.

\bibitem{li2024t2vturbo}
Jiachen Li, Weixi Feng, Tsu-Jui Fu, Xinyi Wang, Sugato Basu, Wenhu Chen, and William~Yang Wang.
\newblock T2v-turbo: Breaking the quality bottleneck of video consistency model with mixed reward feedback.
\newblock {\em arXiv preprint arXiv:2405.18750}, 2024.

\bibitem{li2024controlnet++}
Ming Li, Taojiannan Yang, Huafeng Kuang, Jie Wu, Zhaoning Wang, Xuefeng Xiao, and Chen Chen.
\newblock Controlnet++: Improving conditional controls with efficient consistency feedback: Project page: liming-ai. github. io/controlnet\_plus\_plus.
\newblock In {\em European Conference on Computer Vision}, pages 129--147. Springer, 2024.

\bibitem{liang2022bevfusion}
Tingting Liang, Hongwei Xie, Kaicheng Yu, Zhongyu Xia, Zhiwei Lin, Yongtao Wang, Tao Tang, Bing Wang, and Zhi Tang.
\newblock Bevfusion: A simple and robust lidar-camera fusion framework.
\newblock {\em Advances in Neural Information Processing Systems}, 35:10421--10434, 2022.

\bibitem{lipman2022flow}
Yaron Lipman, Ricky~TQ Chen, Heli Ben-Hamu, Maximilian Nickel, and Matt Le.
\newblock Flow matching for generative modeling.
\newblock {\em arXiv preprint arXiv:2210.02747}, 2022.

\bibitem{liu2022rectified}
Xingchao Liu, Chengyue Gong, and Qiang Liu.
\newblock Flow straight and fast: Learning to generate and transfer data with rectified flow.
\newblock {\em arXiv preprint arXiv:2209.03003}, 2022.

\bibitem{liu2020unstructured}
Yin-Bo Liu, Ming Zeng, and Qing-Hao Meng.
\newblock Unstructured road vanishing point detection using convolutional neural networks and heatmap regression.
\newblock {\em IEEE Transactions on Instrumentation and Measurement}, 70:1--8, 2020.

\bibitem{ma2024unleashing}
Enhui Ma, Lijun Zhou, Tao Tang, Zhan Zhang, Dong Han, Junpeng Jiang, Kun Zhan, Peng Jia, Xianpeng Lang, Haiyang Sun, et~al.
\newblock Unleashing generalization of end-to-end autonomous driving with controllable long video generation.
\newblock {\em arXiv preprint arXiv:2406.01349}, 2024.

\bibitem{mei2024dreamforge}
Jianbiao Mei, Tao Hu, Xuemeng Yang, Licheng Wen, Yu~Yang, Tiantian Wei, Yukai Ma, Min Dou, Botian Shi, and Yong Liu.
\newblock Dreamforge: Motion-aware autoregressive video generation for multi-view driving scenes.
\newblock {\em arXiv preprint arXiv:2409.04003}, 2024.

\bibitem{ni2025wonderturbo}
Chaojun Ni, Xiaofeng Wang, Zheng Zhu, Weijie Wang, Haoyun Li, Guosheng Zhao, Jie Li, Wenkang Qin, Guan Huang, and Wenjun Mei.
\newblock Wonderturbo: Generating interactive 3d world in 0.72 seconds.
\newblock {\em arXiv preprint arXiv:2504.02261}, 2025.

\bibitem{ni2025recondreamerrl}
Chaojun Ni, Guosheng Zhao, Xiaofeng Wang, Zheng Zhu, Wenkang Qin, Xinze Chen, Guanghong Jia, Guan Huang, and Wenjun Mei.
\newblock Recondreamer-rl: Enhancing reinforcement learning via diffusion-based scene reconstruction.
\newblock {\em arXiv preprint arXiv:2508.08170}, 2025.

\bibitem{ni2025recondreamer}
Chaojun Ni, Guosheng Zhao, Xiaofeng Wang, Zheng Zhu, Wenkang Qin, Guan Huang, Chen Liu, Yuyin Chen, Yida Wang, Xueyang Zhang, et~al.
\newblock Recondreamer: Crafting world models for driving scene reconstruction via online restoration.
\newblock In {\em Proceedings of the Computer Vision and Pattern Recognition Conference}, pages 1559--1569, 2025.

\bibitem{nichol2021improved}
Alexander~Quinn Nichol and Prafulla Dhariwal.
\newblock Improved denoising diffusion probabilistic models.
\newblock In {\em International conference on machine learning}, pages 8162--8171. PMLR, 2021.

\bibitem{oquab2023dinov2}
Maxime Oquab, Timothée Darcet, Theo Moutakanni, Huy~V. Vo, Marc Szafraniec, Vasil Khalidov, Pierre Fernandez, Daniel Haziza, Francisco Massa, Alaaeldin El-Nouby, Russell Howes, Po-Yao Huang, Hu~Xu, Vasu Sharma, Shang-Wen Li, Wojciech Galuba, Mike Rabbat, Mido Assran, Nicolas Ballas, Gabriel Synnaeve, Ishan Misra, Herve Jegou, Julien Mairal, Patrick Labatut, Armand Joulin, and Piotr Bojanowski.
\newblock Dinov2: Learning robust visual features without supervision, 2023.

\bibitem{ouyang2022rlhf}
Long Ouyang, Jeffrey Wu, Xu~Jiang, Diogo Almeida, Carroll Wainwright, Pamela Mishkin, Chong Zhang, Sandhini Agarwal, Katarina Slama, Alex Ray, et~al.
\newblock Training language models to follow instructions with human feedback.
\newblock {\em Advances in neural information processing systems}, 35:27730--27744, 2022.

\bibitem{11119091}
Yu~Qian, Xunhao Li, Jian Zhang, Xiaolin Meng, Yongfu Li, Heng Ding, and Maoze Wang.
\newblock A diffusion-tgan framework for spatio-temporal speed imputation and trajectory reconstruction.
\newblock {\em IEEE Transactions on Intelligent Transportation Systems}, pages 1--15, 2025.

\bibitem{11186232}
Yu~Qian, Jian Zhang, Zhanyu Feng, Xunhao Li, Zhiyuan Liu, and Hua Wang.
\newblock Multi-task itransformer: A saturation-based model for short-term highway traffic congestion prediction considering event-induced capacity variability.
\newblock {\em IEEE Transactions on Vehicular Technology}, pages 1--15, 2025.

\bibitem{qiu2023priorlane}
Qibo Qiu, Haiming Gao, Wei Hua, Gang Huang, and Xiaofei He.
\newblock Priorlane: A prior knowledge enhanced lane detection approach based on transformer.
\newblock In {\em 2023 IEEE International Conference on Robotics and Automation (ICRA)}, pages 5618--5624. IEEE, 2023.

\bibitem{rafailov2023dpo}
Rafael Rafailov, Archit Sharma, Eric Mitchell, Christopher~D Manning, Stefano Ermon, and Chelsea Finn.
\newblock Direct preference optimization: Your language model is secretly a reward model.
\newblock {\em Advances in Neural Information Processing Systems}, 36:53728--53741, 2023.

\bibitem{ravi2024sam2}
Nikhila Ravi, Valentin Gabeur, Yuan-Ting Hu, Ronghang Hu, Chaitanya Ryali, Tengyu Ma, Haitham Khedr, Roman R{\"a}dle, Chloe Rolland, Laura Gustafson, Eric Mintun, Junting Pan, Kalyan~Vasudev Alwala, Nicolas Carion, Chao-Yuan Wu, Ross Girshick, Piotr Doll{\'a}r, and Christoph Feichtenhofer.
\newblock Sam 2: Segment anything in images and videos.
\newblock {\em arXiv preprint arXiv:2408.00714}, 2024.

\bibitem{ren2024grounded}
Tianhe Ren, Shilong Liu, Ailing Zeng, Jing Lin, Kunchang Li, He~Cao, Jiayu Chen, Xinyu Huang, Yukang Chen, Feng Yan, Zhaoyang Zeng, Hao Zhang, Feng Li, Jie Yang, Hongyang Li, Qing Jiang, and Lei Zhang.
\newblock Grounded sam: Assembling open-world models for diverse visual tasks, 2024.

\bibitem{rombach2022high}
Robin Rombach, Andreas Blattmann, Dominik Lorenz, Patrick Esser, and Bj{\"o}rn Ommer.
\newblock High-resolution image synthesis with latent diffusion models.
\newblock In {\em Proceedings of the IEEE/CVF conference on computer vision and pattern recognition}, pages 10684--10695, 2022.

\bibitem{schulman2017ppo}
John Schulman, Filip Wolski, Prafulla Dhariwal, Alec Radford, and Oleg Klimov.
\newblock Proximal policy optimization algorithms.
\newblock {\em arXiv preprint arXiv:1707.06347}, 2017.

\bibitem{swerdlow2024bevgen}
Alexander Swerdlow, Runsheng Xu, and Bolei Zhou.
\newblock Street-view image generation from a bird's-eye view layout.
\newblock {\em IEEE Robotics and Automation Letters}, 2024.

\bibitem{tabelini2021laneatt}
Lucas Tabelini, Rodrigo Berriel, Thiago M.~Paix\ ao, Claudine Badue, Alberto Ferreira~De Souza, and Thiago Oliveira-Santos.
\newblock {Keep your Eyes on the Lane: Real-time Attention-guided Lane Detection}.
\newblock In {\em Conference on Computer Vision and Pattern Recognition (CVPR)}, 2021.

\bibitem{tancik2020fourier}
Matthew Tancik, Pratul Srinivasan, Ben Mildenhall, Sara Fridovich-Keil, Nithin Raghavan, Utkarsh Singhal, Ravi Ramamoorthi, Jonathan Barron, and Ren Ng.
\newblock Fourier features let networks learn high frequency functions in low dimensional domains.
\newblock {\em Advances in neural information processing systems}, 33:7537--7547, 2020.

\bibitem{unterthiner2019fvd}
Thomas Unterthiner, Sjoerd Van~Steenkiste, Karol Kurach, Rapha{\"e}l Marinier, Marcin Michalski, and Sylvain Gelly.
\newblock Fvd: A new metric for video generation.
\newblock 2019.

\bibitem{wallace2024diffusiondpo}
Bram Wallace, Meihua Dang, Rafael Rafailov, Linqi Zhou, Aaron Lou, Senthil Purushwalkam, Stefano Ermon, Caiming Xiong, Shafiq Joty, and Nikhil Naik.
\newblock Diffusion model alignment using direct preference optimization.
\newblock In {\em Proceedings of the IEEE/CVF Conference on Computer Vision and Pattern Recognition}, pages 8228--8238, 2024.

\bibitem{wan2025}
Ang Wang, Baole Ai, Bin Wen, Chaojie Mao, Chen-Wei Xie, Di~Chen, Feiwu Yu, Haiming Zhao, Jianxiao Yang, Jianyuan Zeng, Jiayu Wang, Jingfeng Zhang, Jingren Zhou, Jinkai Wang, Jixuan Chen, Kai Zhu, Kang Zhao, Keyu Yan, Lianghua Huang, Mengyang Feng, Ningyi Zhang, Pandeng Li, Pingyu Wu, Ruihang Chu, Ruili Feng, Shiwei Zhang, Siyang Sun, Tao Fang, Tianxing Wang, Tianyi Gui, Tingyu Weng, Tong Shen, Wei Lin, Wei Wang, Wei Wang, Wenmeng Zhou, Wente Wang, Wenting Shen, Wenyuan Yu, Xianzhong Shi, Xiaoming Huang, Xin Xu, Yan Kou, Yangyu Lv, Yifei Li, Yijing Liu, Yiming Wang, Yingya Zhang, Yitong Huang, Yong Li, You Wu, Yu~Liu, Yulin Pan, Yun Zheng, Yuntao Hong, Yupeng Shi, Yutong Feng, Zeyinzi Jiang, Zhen Han, Zhi-Fan Wu, and Ziyu Liu.
\newblock Wan: Open and advanced large-scale video generative models.
\newblock {\em arXiv preprint arXiv:2503.20314}, 2025.

\bibitem{wang2023exploring}
Shihao Wang, Yingfei Liu, Tiancai Wang, Ying Li, and Xiangyu Zhang.
\newblock Exploring object-centric temporal modeling for efficient multi-view 3d object detection.
\newblock {\em arXiv preprint arXiv:2303.11926}, 2023.

\bibitem{wang2024drivedreamer}
Xiaofeng Wang, Zheng Zhu, Guan Huang, Xinze Chen, Jiagang Zhu, and Jiwen Lu.
\newblock Drivedreamer: Towards real-world-drive world models for autonomous driving.
\newblock In {\em European Conference on Computer Vision}, pages 55--72. Springer, 2024.

\bibitem{wang2024scantd}
Yujia Wang, Fang-Lue Zhang, and Neil~A Dodgson.
\newblock Scantd: 360° scanpath prediction based on time-series diffusion.
\newblock In {\em Proceedings of the 32nd ACM International Conference on Multimedia}, pages 7764--7773, 2024.

\bibitem{wang2025target}
Yujia Wang, Fang-Lue Zhang, and Neil~A Dodgson.
\newblock Target scanpath-guided 360-degree image enhancement.
\newblock In {\em Proceedings of the AAAI Conference on Artificial Intelligence}, volume~39, pages 8169--8177, 2025.

\bibitem{wang2024driving}
Yuqi Wang, Jiawei He, Lue Fan, Hongxin Li, Yuntao Chen, and Zhaoxiang Zhang.
\newblock Driving into the future: Multiview visual forecasting and planning with world model for autonomous driving.
\newblock In {\em Proceedings of the IEEE/CVF Conference on Computer Vision and Pattern Recognition}, pages 14749--14759, 2024.

\bibitem{wen2024panacea}
Yuqing Wen, Yucheng Zhao, Yingfei Liu, Fan Jia, Yanhui Wang, Chong Luo, Chi Zhang, Tiancai Wang, Xiaoyan Sun, and Xiangyu Zhang.
\newblock Panacea: Panoramic and controllable video generation for autonomous driving.
\newblock In {\em Proceedings of the IEEE/CVF Conference on Computer Vision and Pattern Recognition}, pages 6902--6912, 2024.

\bibitem{xie2024xdrive}
Yichen Xie, Chenfeng Xu, Chensheng Peng, Shuqi Zhao, Nhat Ho, Alexander~T Pham, Mingyu Ding, Masayoshi Tomizuka, and Wei Zhan.
\newblock X-drive: Cross-modality consistent multi-sensor data synthesis for driving scenarios.
\newblock {\em arXiv preprint arXiv:2411.01123}, 2024.

\bibitem{xin2025lumina}
Yi~Xin, Qi~Qin, Siqi Luo, Kaiwen Zhu, Juncheng Yan, Yan Tai, Jiayi Lei, Yuewen Cao, Keqi Wang, Yibin Wang, et~al.
\newblock Lumina-dimoo: An omni diffusion large language model for multi-modal generation and understanding.
\newblock {\em arXiv preprint arXiv:2510.06308}, 2025.

\bibitem{xu2023imagereward}
Jiazheng Xu, Xiao Liu, Yuchen Wu, Yuxuan Tong, Qinkai Li, Ming Ding, Jie Tang, and Yuxiao Dong.
\newblock Imagereward: Learning and evaluating human preferences for text-to-image generation.
\newblock {\em Advances in Neural Information Processing Systems}, 36:15903--15935, 2023.

\bibitem{yan2024drivingsphere}
Tianyi Yan, Dongming Wu, Wencheng Han, Junpeng Jiang, Xia Zhou, Kun Zhan, Cheng-zhong Xu, and Jianbing Shen.
\newblock Drivingsphere: Building a high-fidelity 4d world for closed-loop simulation.
\newblock {\em arXiv preprint arXiv:2411.11252}, 2024.

\bibitem{yan2025olidm}
Tianyi Yan, Junbo Yin, Xianpeng Lang, Ruigang Yang, Cheng-Zhong Xu, and Jianbing Shen.
\newblock Olidm: Object-aware lidar diffusion models for autonomous driving.
\newblock In {\em Proceedings of the AAAI Conference on Artificial Intelligence}, volume~39, pages 9121--9129, 2025.

\bibitem{yang2024using}
Kai Yang, Jian Tao, Jiafei Lyu, Chunjiang Ge, Jiaxin Chen, Weihan Shen, Xiaolong Zhu, and Xiu Li.
\newblock Using human feedback to fine-tune diffusion models without any reward model.
\newblock In {\em Proceedings of the IEEE/CVF Conference on Computer Vision and Pattern Recognition}, pages 8941--8951, 2024.

\bibitem{yang2023bevcontrol}
Kairui Yang, Enhui Ma, Jibin Peng, Qing Guo, Di~Lin, and Kaicheng Yu.
\newblock Bevcontrol: Accurately controlling street-view elements with multi-perspective consistency via bev sketch layout.
\newblock {\em arXiv preprint arXiv:2308.01661}, 2023.

\bibitem{depth_anything_v2}
Lihe Yang, Bingyi Kang, Zilong Huang, Zhen Zhao, Xiaogang Xu, Jiashi Feng, and Hengshuang Zhao.
\newblock Depth anything v2.
\newblock {\em arXiv:2406.09414}, 2024.

\bibitem{yang2025ipo}
Xiaomeng Yang, Zhiyu Tan, and Hao Li.
\newblock Ipo: Iterative preference optimization for text-to-video generation.
\newblock {\em arXiv preprint arXiv:2502.02088}, 2025.

\bibitem{yang2024drivearena}
Xuemeng Yang, Licheng Wen, Yukai Ma, Jianbiao Mei, Xin Li, Tiantian Wei, Wenjie Lei, Daocheng Fu, Pinlong Cai, Min Dou, et~al.
\newblock Drivearena: A closed-loop generative simulation platform for autonomous driving.
\newblock {\em arXiv preprint arXiv:2408.00415}, 2024.

\bibitem{yang2025matrix}
Zhongqi Yang, Wenhang Ge, Yuqi Li, Jiaqi Chen, Haoyuan Li, Mengyin An, Fei Kang, Hua Xue, Baixin Xu, Yuyang Yin, et~al.
\newblock Matrix-3d: Omnidirectional explorable 3d world generation.
\newblock {\em arXiv preprint arXiv:2508.08086}, 2025.

\bibitem{yi2024towards}
Mingyang Yi, Aoxue Li, Yi~Xin, and Zhenguo Li.
\newblock Towards understanding the working mechanism of text-to-image diffusion model.
\newblock {\em Advances in Neural Information Processing Systems}, 37:55342--55369, 2024.

\bibitem{yu2023flashocc}
Zichen Yu, Changyong Shu, Jiajun Deng, Kangjie Lu, Zongdai Liu, Jiangyong Yu, Dawei Yang, Hui Li, and Yan Chen.
\newblock Flashocc: Fast and memory-efficient occupancy prediction via channel-to-height plugin.
\newblock 2023.

\bibitem{yuan2024instructvideo}
Hangjie Yuan, Shiwei Zhang, Xiang Wang, Yujie Wei, Tao Feng, Yining Pan, Yingya Zhang, Ziwei Liu, Samuel Albanie, and Dong Ni.
\newblock Instructvideo: Instructing video diffusion models with human feedback.
\newblock In {\em Proceedings of the IEEE/CVF Conference on Computer Vision and Pattern Recognition}, pages 6463--6474, 2024.

\bibitem{zeng2024driving}
Shuang Zeng, Xinyuan Chang, Xinran Liu, Zheng Pan, and Xing Wei.
\newblock Driving with prior maps: Unified vector prior encoding for autonomous vehicle mapping.
\newblock {\em arXiv preprint arXiv:2409.05352}, 2024.

\bibitem{zeng2025FSDrive}
Shuang Zeng, Xinyuan Chang, Mengwei Xie, Xinran Liu, Yifan Bai, Zheng Pan, Mu~Xu, and Xing Wei.
\newblock Futuresightdrive: Thinking visually with spatio-temporal cot for autonomous driving.
\newblock {\em arXiv preprint arXiv:2505.17685}, 2025.

\bibitem{zhang2024onlinevpo}
Jiacheng Zhang, Jie Wu, Weifeng Chen, Yatai Ji, Xuefeng Xiao, Weilin Huang, and Kai Han.
\newblock Onlinevpo: Align video diffusion model with online video-centric preference optimization.
\newblock {\em arXiv preprint arXiv:2412.15159}, 2024.

\bibitem{rongchao}
Rongchao Zhang, Yu~Huang, Yiwei Lou, Weiping Ding, Yongzhi Cao, and Hanpin Wang.
\newblock Synergistic attention-guided cascaded graph diffusion model for complementarity determining region synthesis.
\newblock {\em {IEEE} Trans. Neural Networks Learn. Syst.}, 36(7):11875--11886, 2025.

\bibitem{zhao2025drivedreamer2}
Guosheng Zhao, Xiaofeng Wang, Zheng Zhu, Xinze Chen, Guan Huang, Xiaoyi Bao, and Xingang Wang.
\newblock Drivedreamer-2: Llm-enhanced world models for diverse driving video generation.
\newblock In {\em Proceedings of the AAAI Conference on Artificial Intelligence}, volume~39, pages 10412--10420, 2025.

\bibitem{zheng2024opensora}
Zangwei Zheng, Xiangyu Peng, Tianji Yang, Chenhui Shen, Shenggui Li, Hongxin Liu, Yukun Zhou, Tianyi Li, and Yang You.
\newblock Open-sora: Democratizing efficient video production for all.
\newblock {\em arXiv preprint arXiv:2412.20404}, 2024.

\end{thebibliography}
\newpage
\section*{NeurIPS Paper Checklist}

\begin{enumerate}

\item {\bf Claims}
    \item[] Question: Do the main claims made in the abstract and introduction accurately reflect the paper's contributions and scope?
    \item[] Answer: \answerYes{} % Replace by \answerYes{}, \answerNo{}, or \answerNA{}.
    \item[] Justification: 

\item {\bf Limitations}
    \item[] Question: Does the paper discuss the limitations of the work performed by the authors?
    \item[] Answer: \answerYes{} % Replace by \answerYes{}, \answerNo{}, or \answerNA{}.
    \item[] Justification: In the  supplementary materials

\item {\bf Theory assumptions and proofs}
    \item[] Question: For each theoretical result, does the paper provide the full set of assumptions and a complete (and correct) proof?
    \item[] Answer: \answerNA{} % Replace by \answerYes{}, \answerNo{}, or \answerNA{}.
    \item[] Justification: not include theoretical results

    \item {\bf Experimental result reproducibility}
    \item[] Question: Does the paper fully disclose all the information needed to reproduce the main experimental results of the paper to the extent that it affects the main claims and/or conclusions of the paper (regardless of whether the code and data are provided or not)?
    \item[] Answer: \answerYes{} % Replace by \answerYes{}, \answerNo{}, or \answerNA{}.
    \item[] Justification: Detailed implementation details are included in the supplementary materials.

\item {\bf Open access to data and code}
    \item[] Question: Does the paper provide open access to the data and code, with sufficient instructions to faithfully reproduce the main experimental results, as described in supplemental material?
    \item[] Answer: \answerNo{} % Replace by \answerYes{}, \answerNo{}, or \answerNA{}.
    \item[] Justification: We will open source the code once we are ready.

\item {\bf Experimental setting/details}
    \item[] Question: Does the paper specify all the training and test details (e.g., data splits, hyperparameters, how they were chosen, type of optimizer, etc.) necessary to understand the results?
    \item[] Answer: \answerYes{} % Replace by \answerYes{}, \answerNo{}, or \answerNA{}.
    \item[] Justification: Details will be included in the included in the supplementary materials.

\item {\bf Experiment statistical significance}
    \item[] Question: Does the paper report error bars suitably and correctly defined or other appropriate information about the statistical significance of the experiments?
    \item[] Answer: \answerNo{} % Replace by \answerYes{}, \answerNo{}, or \answerNA{}.
    \item[] Justification: 

\item {\bf Experiments compute resources}
    \item[] Question: For each experiment, does the paper provide sufficient information on the computer resources (type of compute workers, memory, time of execution) needed to reproduce the experiments?
    \item[] Answer: \answerYes{} % Replace by \answerYes{}, \answerNo{}, or \answerNA{}.
    \item[] Justification: 
    
\item {\bf Code of ethics}
    \item[] Question: Does the research conducted in the paper conform, in every respect, with the NeurIPS Code of Ethics \url{https://neurips.cc/public/EthicsGuidelines}?
    \item[] Answer: \answerYes{} % Replace by \answerYes{}, \answerNo{}, or \answerNA{}.
    \item[] Justification: 

\item {\bf Broader impacts}
    \item[] Question: Does the paper discuss both potential positive societal impacts and negative societal impacts of the work performed?
    \item[] Answer: \answerNA{} % Replace by \answerYes{}, \answerNo{}, or \answerNA{}.
    \item[] Justification: 
    
\item {\bf Safeguards}
    \item[] Question: Does the paper describe safeguards that have been put in place for responsible release of data or models that have a high risk for misuse (e.g., pretrained language models, image generators, or scraped datasets)?
    \item[] Answer: \answerNA{} % Replace by \answerYes{}, \answerNo{}, or \answerNA{}.
    \item[] Justification:

\item {\bf Licenses for existing assets}
    \item[] Question: Are the creators or original owners of assets (e.g., code, data, models), used in the paper, properly credited and are the license and terms of use explicitly mentioned and properly respected?
    \item[] Answer: \answerNA{}% Replace by \answerYes{}, \answerNo{}, or \answerNA{}.
    \item[] Justification: 

\item {\bf New assets}
    \item[] Question: Are new assets introduced in the paper well documented and is the documentation provided alongside the assets?
    \item[] Answer: \answerNA{} % Replace by \answerYes{}, \answerNo{}, or \answerNA{}.
    \item[] Justification: 

\item {\bf Crowdsourcing and research with human subjects}
    \item[] Question: For crowdsourcing experiments and research with human subjects, does the paper include the full text of instructions given to participants and screenshots, if applicable, as well as details about compensation (if any)? 
    \item[] Answer:\answerNA{} % Replace by \answerYes{}, \answerNo{}, or \answerNA{}.
    \item[] Justification: 

\item {\bf Institutional review board (IRB) approvals or equivalent for research with human subjects}
    \item[] Question: Does the paper describe potential risks incurred by study participants, whether such risks were disclosed to the subjects, and whether Institutional Review Board (IRB) approvals (or an equivalent approval/review based on the requirements of your country or institution) were obtained?
    \item[] Answer: \answerNA{} % Replace by \answerYes{}, \answerNo{}, or \answerNA{}.
    \item[] Justification: 

\item {\bf Declaration of LLM usage}
    \item[] Question: Does the paper describe the usage of LLMs if it is an important, original, or non-standard component of the core methods in this research? Note that if the LLM is used only for writing, editing, or formatting purposes and does not impact the core methodology, scientific rigorousness, or originality of the research, declaration is not required.
    %this research? 
    \item[] Answer: \answerNo{} % Replace by \answerYes{}, \answerNo{}, or \answerNA{}.
    \item[] Justification: Only for editing.

\end{enumerate}

\end{document}